\documentclass[twoside,11pt]{article}

\usepackage[abbrvbib, preprint]{jmlr2e}
\usepackage{amsmath,amssymb,amsfonts}
\usepackage{algorithm} 
\usepackage{algpseudocode} 
\usepackage{graphicx}
\usepackage{tabularx}
\usepackage{makecell}
\usepackage{multirow}
\usepackage{textcomp}
\usepackage[table]{xcolor}
\usepackage{xcolor}
\usepackage{caption}
\usepackage{subcaption}
\usepackage{dirtytalk}
\usepackage{epsdice}
\usepackage{rotating}
\usepackage{tikz}
\usepackage{dutchcal}
\usepackage{booktabs}
\usepackage{adjustbox}
\usepackage{bm}

\usetikzlibrary{arrows,shapes,positioning,shadows,trees, 3d, calc, shapes.misc, patterns}
\def\BibTeX{{\rm B\kern-.05em{\sc i\kern-.025em b}\kern-.08em
    T\kern-.1667em\lower.7ex\hbox{E}\kern-.125emX}}

\definecolor{input_purple}{HTML}{431cce} 
\definecolor{middle_beige}{HTML}{FAE7D6}
\definecolor{output_red}{HTML}{CE1C4E}
\definecolor{text_white}{HTML}{ECECEC}

\tikzset{
    in_out/.style = {draw, thick, rectangle, minimum height = 0.6cm,  minimum width = 1.2cm, rounded corners=5, align=center, fill=middle_beige},
    archi_layer/.style = {draw, thick, circle, align=center, fill=middle_beige},
    a_dia/.style = {draw, thick, diamond, align=center, fill=middle_beige},
    input/.style = {coordinate},
    num_circle/.style = {draw, thick, circle, black, fill=middle_beige}}

\DeclareMathOperator*{\argmin}{arg\,min}

\author{\IEEEauthorblockN{Julie Keisler}
\IEEEauthorblockA{\textit{EDF Lab Paris-Saclay} \\
\textit{University of Lille \& INRIA}\\
julie.keisler@edf.fr}
\and
\IEEEauthorblockN{El-Ghazali Talbi}
\IEEEauthorblockA{\textit{University of Lille \& INRIA} \\
el-ghazali.talbi@univ-lille.fr}
\and
\IEEEauthorblockN{Sandra Claudel}
\IEEEauthorblockA{\textit{EDF Lab Paris-Saclay} \\
sandra.claudel@edf.fr}
\and
\IEEEauthorblockN{Gilles Cabriel}
\IEEEauthorblockA{\textit{EDF Lab Paris-Saclay} \\
gilles.cabriel@edf.fr}
}
\usepackage{blindtext}

\usepackage{lastpage}
\jmlrheading{None}{None}{1-\pageref{LastPage}}{None; Revised None}{None}{None}{Julie Keisler, El-Ghazali Talbi, Sandra Claudel and Gilles Cabriel}

\ShortHeadings{Optimization framework for AutoDL}{Optimization framework for AutoDL}
\firstpageno{1}

\begin{document}

\title{An algorithmic framework for the optimization of deep neural networks architectures and hyperparameters}

\author{\name Julie Keisler \email julie.keisler@edf.fr \\
       \addr EDF Lab Paris-Saclay\\
       Bd Gaspard Monge, 91120 Palaiseau\\
       University of Lille \& INRIA\\
       170 Av. de Bretagne, 59000 Lille
       \AND
       \name El-Ghazali Talbi \email el-ghazali.talbi@univ-lille.fr \\
       \addr University of Lille \& INRIA\\
       170 Av. de Bretagne, 59000 Lille
       \AND
       \name Sandra Claudel \email sandra.claudel@edf.fr \\
       \addr EDF Lab Paris-Saclay\\
       Bd Gaspard Monge, 91120 Palaiseau
       \AND
       \name Gilles Cabriel \email gilles.cabriel@edf.fr \\
       \addr EDF Lab Paris-Saclay\\
       Bd Gaspard Monge, 91120 Palaiseau}

\editor{My editor}

\maketitle

\begin{abstract}
In this paper, we propose DRAGON (for DiRected Acyclic Graph OptimizatioN), an algorithmic framework to automatically generate efficient deep neural networks and optimize their associated hyperparameters. The framework is based on evolving directed acyclic graphs (DAGs), defining a more flexible search space than the existing ones in the literature. It allows mixtures of different classical operations: convolutions, recurrences and dense layers, but also more newfangled operations such as self-attention. Based on this search space we propose neighbourhood and evolution search operators to optimize both the architecture and hyper-parameters of our networks. These search operators can be used with any metaheuristic capable of handling mixed search spaces. We tested our algorithmic framework with an asynchronous evolutionary algorithm on a time series forecasting benchmark. The results demonstrate that DRAGON outperforms state-of-the-art handcrafted models and AutoML techniques for time series forecasting on numerous datasets. DRAGON has been implemented as a python open-source package\footnote{https://dragon-tutorial.readthedocs.io/en/latest/index.html}.

\end{abstract}

\begin{keywords}
Metaheuristics, Evolutionary Algorithm, Neural Architecture Search, Hyperparameter optimization, Time Series Forecasting
\end{keywords}

\section{Introduction}

With the recent successes of deep learning in many research fields, deep neural networks (DNN) optimization stimulates the growing interest of the scientific community \citep{talbi2021automated}. While each new learning task requires the handcrafted design of a new DNN, automated deep learning facilitates the creation of powerful DNNs. Interests are to give access to deep learning to less experienced people, to reduce the tedious tasks of managing many parameters to reach the optimal DNN, and finally, to go beyond what humans can design by creating non-intuitive DNNs that can ultimately prove to be more efficient.

Optimizing a DNN means automatically finding an optimal architecture for a given learning task: choosing the operations and the connections between those operations and the associated hyperparameters. The first task is also known as Neural Architecture Search \citep{elsken2019neural}, also named NAS, and the second, as HyperParameters Optimization (HPO). Most works from the literature try to tackle only one of these two optimization problems. Many papers related to NAS \citep{white2021bananas,loni_deepmaker_2020, wang_evolving_2019, sun_particle_2018, zhong_dna_2020} focus on designing optimal architectures for computer vision tasks with stacked convolution and pooling layers. Because each DNN training is time-consuming, researchers tried to reduce the search space by adding many constraints preventing from finding irrelevant architectures. These strategies are relevant in the case of computer vision or NLP, where the models to be trained are huge and the high performance architectures are well identified. However, there is a gap in the literature regarding the use of NAS and HPO for problems where neural networks could be efficient, but the relevant models have not been clearly identified.

To fill this gap, we introduce DRAGON (for DiRected Acyclic Graphs OptimizatioN), a new optimization framework for DNNs based on the evolution of Directed Acyclic Graphs (DAGs). The encoding and the search operators are higlhy flexible and may be used with various deep learning and AutoML problems. We ran experiments on time series forecasting tasks and demonstrate on a large variety of datasets that DRAGON can find DNNs which outperform state-of-the-art handcrafted forecasters and AutoML frameworks. In summary, our contributions are as follows:

\begin{itemize}
\item The precise definition of a flexible search space based on DAGs, for the optimization of DNN architectures and hyperparameters. This search space may be used for various tasks, and is particularly useful when the performing architectures for a given problem are not clearly identified.

\item The design of efficient neighbourhoods and variation operators for DAGs. With these operators, any metaheuristic designed for a mixed and variable-size search space can be applied. In this paper, we investigate the use of an asynchronous evolutionary algorithm.

\item The validation of the algorithmic framework on a popular time series forecasting benchmark \citep{godahewa2021monash}. We compare ourselves with 15 handcrafted statistical and machine learning models \citep{godahewa2021monash} as well as 6 AutoML frameworks on 27 datasets \citep{shchur2023autogluon}. We show that DRAGON outperforms the 21 models from this baseline on 11 out of 27 datasets. The only competitive model is the AutoML framework AutoGluon \citep{shchur2023autogluon}, which outperforms the baseline on 10 out of 27 datasets and was beaten by DRAGON on 14 out of 27 datasets.
\end{itemize}

The paper is organized as follows: we review section~\ref{part 2}, the literature on deep learning models for time series forecasting and AutoML. Section~\ref{part 3} defines our search space. Section~\ref{part 4} presents our neighbourhoods and variation operators within the evolutionary algorithm. Section~\ref{part5} details our experimental results obtained on a popular time series forecasting benchmark. Finally, Section~\ref{part 6} gives a conclusion and introduces further research opportunities.
\section{Related Work} \label{part 2}

\subsection{Deep learning for time series forecasting}

\label{part dl related}

Time series forecasting has been studied for decades. The field has been dominated for a long time by statistical tools such as ARIMA, Exponential Smoothing (ES), or (S)ARIMAX, this last model allowing the use of exogenous variables. It now opens itself to deep learning models \citep{9461796}. These new models recently achieved great performances on many datasets. Three main parts compose typical DNNs: an input layer, several hidden layers and an output layer. In this paper we apply our framework to optimize the hidden layers for a given time series forecasting task (see Figure~\ref{fig:metamodel_monash}). In this part, we introduce usual DNN layers for time series forecasting, which can be used in our search space.

The first layer type from our search space is the fully-connected layer, or Multi-Layer Perceptron (MLP). The input vector is multiplied by a weight matrix. Most architectures use such layers as simple building blocks for dimension matching, input embedding or output modelling. The N-Beats model is a well-known example of a DNN based on fully-connected layers for time series forecasting \citep{NBeats}.

The second layer type \citep{lecun2015deep} is the convolution layer (CNN). Inspired by the human brain's visual cortex, it has mainly been popularised for computer vision. The convolution layer uses a discrete convolution operator between the input data and a small matrix called a filter. The extracted features are local and time-invariant if the considered data are time series. Many architectures designed for time series forecasting are based on convolution layers such as WaveNet \citep{oord2016wavenet} and Temporal Convolution Networks \citep{lea2017temporal}.

The third layer type is the recurrent layer (RNN), specifically designed for sequential data processing, therefore, particularly suitable for time series. These layers scan the sequential data and keep information from the sequence past in memory to predict its future. A popular model based on RNN layers is the Seq2Seq network \citep{seq2seq}. Two RNNs, an encoder and a decoder, are sequentially connected by a fixed-length vector. Various versions of the Seq2Seq model have been introduced in the literature, such as the DeepAR model \citep{salinas2020deepar}, which encompasses an RNN encoder in an autoregressive model. The major weakness of RNN layers is the modelling of long-term dynamics due to the vanishing gradient. Long Short-Term Memory (LSTM) and Gated Recurrent Unit (GRU) layers have been introduced \citep{hochreiter1997long, chung2014empirical} to overcome this problem.

Finally, the layer type from our search space is the attention layer. The attention layer has been popularized within the deep learning community as part of Vaswani's transformer model \citep{vaswani2017attention}. The attention layer is more generic than the convolution. It can model the dependencies of each element from the input sequence with all the others. In the vanilla transformer \citep{vaswani2017attention}, the attention layer does not factor the relative distance between inputs in its modelling but rather the element's absolute position in the sequence. The Transformer-XL \citep{dai2019transformer}, a transformer variant created to tackle long-term dependencies tasks, introduces a self-attention version with relative positions. \citet{cordonnier2019relationship} used this new attention formulation to show that, under a specific configuration of parameters, the attention layers could be trained as convolution layers. Within our search space, we chose this last formulation of attention, with the relative positions.

The three first layers (i.e. MLP, CNN, RNN) were frequently mixed into DNN architectures. Sequential and parallel combinations of convolution, recurrent and fully connected layers often compose state-of-the-art DNN models for time series forecasting. Layer diversity enables the extraction of different and complementary features from input data to allow a better prediction. Some recent DNN models introduce transformers into hybrid DNNs. In \citet{lim2021temporal}, the authors developed the Temporal Fusion Transformer, a hybrid model stacking transformer layers on top of an RNN layer. With this in mind, we built a flexible search space which generalizes hybrid DNN models including MLPs, CNNs, RNNs and transformers.

\subsection{Search spaces for automated deep learning}

Designing an efficient DNN for a given task requires choosing an architecture and tuning its many hyperparameters. It is a difficult, fastidious, and time-consuming optimization task. Moreover, it requires expertise and restricts the discovery of new DNNs to what humans can design. Research related to the automatic design and optimization of DNNs has therefore risen this last decade \citep{talbi2021automated}. The first challenge in automatic deep learning (AutoDL), and more specifically neural architecture search (NAS), is search space design. Typical search spaces for Hyperparameters Optimization (HPO) are a product space of a mixture of continuous and categorical dimensions (e.g. learning rate, number of layers, batch size), while NAS focuses on optimizing the topology of the DNN \citep{white2023neural}. Encoding a DNN topology is a complex task because the encoding should not be too broad and allow too many architectures to keep the search efficient. On the contrary, if the encoding is too restrictive, we may miss promising solutions and novel architectures. This means before creating the search space we need to choose which DNNs or type of DNNs are relevant or not to the problem at hand. Once we have decided on this broad set of DNNs, we define the search space following a set of rules \citep{talbi2021automated}:

\begin{itemize}
\item Completeness: all (or almost all) relevant DNNs from this broad set should be encoded in the search space.
\item Connectedness: a path should always be possible between two encoded DNNs in the search space.
\item Efficiency: the encoding should be easy to manipulate by the search operators (i.e. neighbourhoods, variation operators) of the search strategy.
\item Constraint handling: the encoding should facilitate the handling of the various constraints to generate feasible DNNs.
\end{itemize}

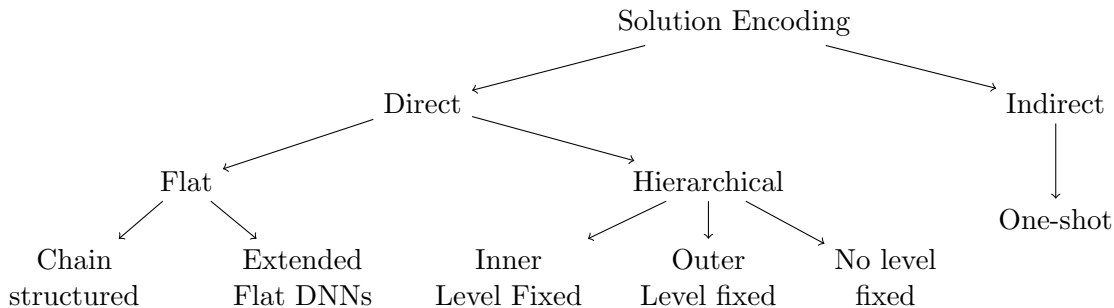
\begin{figure*}[htbp]
\centering
    \begin{tikzpicture}
        \draw node[] (SE) {Solution Encoding};
        \draw node[below left =0.5cm and 1.8cm of SE] (Direct) {Direct};
        \draw node[below right =0.5cm and 1.8cm of SE] (Indirect) {Indirect};
        \draw node[below left =0.5cm and 2cm of Direct] (Flat) {Flat};
        \draw node[below right =0.5cm and 2cm of Direct] (Hierarchical) {Hierarchical};
        \draw node[below left =0.5cm and 0cm of Flat, align=center] (Chain) {Chain\\structured};
        \draw node[below right =0.5cm and 0cm of Flat, align=center] (Extended) {Extended\\Flat DNNs};
        \draw node[below =0.5cm of Hierarchical, align=center] (Outer) {Outer\\Level fixed};
        \draw node[left =0.5cm of Outer, align=center] (Inner) {Inner\\Level Fixed};
        \draw node[right =0.5cm of Outer, align=center] (No) {No level\\fixed};
        \draw node[below =of Indirect, align=center] (One) {One-shot};
        \draw[->] (SE) -- (Direct);
        \draw[->] (SE) -- (Indirect);
        \draw[->] (Direct) -- (Flat);
        \draw[->] (Direct) -- (Hierarchical);
        \draw[->] (Flat) -- (Chain);
        \draw[->] (Flat) -- (Extended);
        \draw[->] (Hierarchical) -- (Inner);
        \draw[->] (Hierarchical) -- (Outer);
        \draw[->] (Hierarchical) -- (No);
        \draw[->] (Indirect) -- (One);
    \end{tikzpicture}
    \caption{Classification of encoding strategies for NAS \citep{talbi2021automated}.}
    \label{fig:classi_encoding}
\end{figure*}

A complete classification of encoding strategies for NAS is presented in \citet{talbi2021automated} and reproduced in Figure \ref{fig:classi_encoding}. We can discriminate between direct and indirect encodings. With direct strategies, the DNNs are completely defined by the encoding, while indirect strategies need a decoder to find the architecture back. Amongst direct strategies, one can discriminate between two categories: flat and hierarchical encodings. In flat encodings, all layers are individually encoded \citep{loni2020deepmaker, sun2018particle, wang2018evolving, wang2019evolving}. The global architecture can be a single chain, with each layer having a single input and a single output, which is called chain structured \citep{assuncao_denser_2018}, but more complex patterns such as multiple outputs, skip connections, have been introduced in the extended flat DNNs encoding \citep{chen_scale-aware_2021}. For hierarchical encodings, they are bundled in blocks \citep{pham2018efficient, shu2019understanding, liu2017hierarchical, zhang2019d}. If the optimization is made on the sequencing of the blocks, with an already chosen content, this is referred to as inner-level fixed \citep{camero2021bayesian, white2021bananas}. If the optimization is made on the blocks' content with a fixed sequencing, it is called outer level fixed. A joint optimization with no level fixed is also an option \citep{liu2019auto}. Regarding the indirect strategies, one popular encoding is the one-shot architecture \citep{bender2018understanding, brock2017smash}. One single large network resuming all candidates from the search space is trained. Then the architectures are found by pruning some branches. Only the best promising architectures are retrained from scratch.

Our search space can be categorized as a direct and extended flat encoding. It is based on the representation of DNNs by DAGs. This representation is very popular among the NAS community and is used by cell-based search spaces such as NAS-Bench-101 inspired by the ResNet architecture \citep{ying2019bench}, as well as one-shot representation such as the DARTS framework (for Differentiable Architecture Search) proposed by \citet{liu2018darts}. In cell-based search spaces, DNNs are represented by repeated cells encoded as DAGs, where each node is an operation belonging to a well-defined list, typically: convolution of size 1, 3, or 5, pooling of size 3, skip connection, or zeroed operation for an image classification task for example. The graphs are then represented either as vectors using path encoding, or as adjacency matrices. In the case of path encoding, different search algorithms can be used, such as Bayesian optimization \citep{white_bananas_2020}, reinforcement learning \citep{zoph2018learning}, particle swarm optimization \citep{wang_evolving_2019}, or evolutionary algorithms \citep{xie2017genetic}, for which classical mutation and crossover operators are usually used and consist in modifying the elements of the path. Adjacency matrices, on the other hand, are more complex objects to optimize. The matrix itself represents the connections within the graph and is usually accompanied by a list representing the nodes content. In the literature, these matrices have been optimized directly with random search algorithms \citep{irwin2019graph} or indirectly with neural predictors based on auto-encoders (see for example \citet{zhang2019d} or \citet{chatzianastasis2021graph}). In the case of one-shot representations, an initial large graph containing all the considered DNN is pruned with a certain search algorithm to only keep the best possible subgraph (and thus the best possible DNN). Various search algorithms can be used to simplify this meta-graph \citep{bender2018understanding} like evolutionary algorithm \citep{guo2020single}. One of the most widely used techniques is DARTS \citep{liu2018darts}, where each edge is associated with a candidate operation, assigned to a probability of being retained in the final subgraph, optimized by gradient descent. The candidate operations can be very traditional, such as for cell-based search spaces, but \citet{chen_scale-aware_2021} proposes to use DARTS with other types of operations, such as inter-variable attention, for multivariate time series prediction. While such search spaces have proven to be efficient for tasks like image classification or language processing, \citet{white2023neural} points out that current NAS search spaces are not very expressive and prevents finding highly novel architectures. This problem is amplified when dealing with tasks for which no known architectures have yet been found.

 Compared to these search spaces, the one we define in this paper is more flexible. We address the optimization of both the architecture and the hyperparameters. We do not fix a list of possible operations with fixed hyperparameters, as is done in these works, but leave the user free to use any operation coded as \textit{PyTorch nn.Module} and to optimize any chosen parameters. Furthermore, we do not fix the generic form of our graph, we do not fix a maximum number of incoming or outgoing edges and we allow to expand or reduce the graphs. DRAGON is capable of generating innovative, original, yet well-performing DNNs. This flexibility may hinder the framework's ability to find good DNNs compared to the NAS state-of-the-art for well-known tasks such as image classification or language processing. However, in cases where DNNs have not been extensively studied and well-performing architectures have not yet been found, such as time series prediction, DRAGON may be more useful and powerful. Finally, we encode our DAGs using their adjacency matrices and provide evolutionary operators to directly modify this representation. To our knowledge, neither such a large search space nor such operators have been used in the literature.

\subsection{AutoML for time series forecasting}

The automated design of DNNs called Automated Deep Learning (AutoDL), belongs to a larger field \citep{hutter2019automated} called Automated Machine Learning (AutoML). AutoML aims to automatically design well-performing machine learning pipelines, for a given task. Works on model optimization for time series forecasting mainly focused on AutoML rather than AutoDL \citep{alsharef2022review}. The optimization can be performed at several levels: input features selection, extraction and engineering, model selection and hyperparameters tuning. Initial research works used to focus on one of these subproblems, while more recent works offer complete optimization pipelines.

The first subproblems, input features selection, extraction and engineering, are specific to our learning task: time series forecasting. This tedious task can significantly improve the prediction scores by giving the model relevant information about the data. Methods to select the features are among computing the importance of each feature on the results or using statistical tools on the signals to extract relevant information. Next, the model selection aims at choosing among a set of diverse machine learning models the best-performing one on a given task. Often, the models are trained separately, and the best model is chosen. In general, the selected model has many hyperparameters, such as the number of hidden layers, activation function or learning rate. Their optimization usually allows for improving the performance of the model.

Nowadays, many research works implement complete optimization pipelines combining those subproblems for time series forecasting. The Time Series Pipeline Optimization framework \citep{dahl2020tspo}, is based on an evolutionary algorithm to automatically find the right features thanks to input signal analysis, then the model and its related hyperparameters. AutoAI-TS \citep{shah2021autoai} is also a complete optimization pipeline, with model selection performed among a wide assortment of models: statistical models, machine learning, deep learning models and hybrids models. Closer to our work, the framework Auto-Pytorch-TS \citep{deng2022efficient} is specific to deep learning models optimization for time series forecasting. The framework uses Bayesian optimization with multi-fidelity optimization. Finally, a recent work from Amazon \citep{shchur2023autogluon} introduces a time series version to their AutoML framework, AutoGluon, leveraging ensembles of statistical and machine learning forecasters.

Except for AutoPytorch-TS, cited works covering the entire optimization pipeline for time series do not deepen model optimization and only perform model selection and hyperparameters optimization. However, time series data becomes more complex, and there is a growing need for more sophisticated and data-specific DNNs. Our framework DRAGON, presented in this paper, only tackles the model selection and hyperparameters optimization parts of the pipeline. We made this choice to show the effectiveness of our framework for designing better DNNs. If we had implemented feature selection, it would have been harder to determine whether the superiority of our results came from the input features pool or the model itself.
\section{Search space definition} \label{part 3}

The development of our optimization framework DRAGON requires the definition of a search space, an objective function and a search algorithm. In this section, we formulate the handled optimization problem an then we detail DRAGON's search space and its characteristics.

\subsection{Optimization problem formulation}

Our optimization problem consists in finding the best possible DNN for a given time series forecasting problem. To do so, we introduce an ensemble $\Omega$ representing our search space, which contains all considered DNNs. We then consider our time series dataset $\mathcal{D}$. For any subset $\mathcal{D}_0 = (X_0, Y_0) \in \mathcal{P}(\mathcal{D})$, we define the forecast error $\ell$ as:
\begin{align*}
  \ell \colon  \Omega \times \mathcal{P}(\mathcal{D}) &\to \mathbb{R} \\
    f \times \mathcal{D}_0 &\mapsto \ell\big(f(\mathcal{D}_0)\big) = \ell\big(Y_0, f(X_0)\big) \, . 
\end{align*}

The explicit formula for $\ell$ will be given later in the paper. Each element $f$ from $\Omega$ is a DNN defined as an operator parameterized by three parameters. First, its architecture $\alpha~\in~\mathcal{A}$. The search space of all considered architectures is called $\mathcal{A}$ and will be detailed in Subsection~\ref{part sp archi}. Given the DNN architecture $\alpha$, the DNN is then parameterized by its hyperparameters $\lambda \in {\Lambda(\alpha)}$, with $\Lambda(\alpha)$ the search space of the hyperparameters induced by the architecture $\alpha$ and defined Subsection~\ref{part sp hp}. Finally, $\alpha$ and $\lambda$ generate an ensemble of possible weights $\Theta(\alpha, \lambda)$, from which the DNN optimal weights $\theta$ are found by gradient descent when training the model. The architecture $\alpha$ and the hyperparameters $\lambda$ are optimized by our framework DRAGON.

We consider the multivariate time series forecasting task. Our dataset $\mathcal{D} = (X, Y)$ is composed of a target variable $Y = \{\mathbf{y}_t\}_{t=1}^{T}$, with $\mathbf{y}_t \in \mathbb{R}^{N}$ the target value at the time step $t$, and a set of explanatory variables (features) $X = \{\mathbf{x}_t\}_{t=1}^{T}$, with $\mathbf{x}_t \in \mathbb{R}^{F_1 \times F_2}$. The size of the target $Y$ at each time step is $N$ and $F_1$, $F_2$ are the shapes of the input variable $X$ at each time step. We choose to represent $\mathbf{x}_t$ by a matrix to extend our framework's scope, but it can equally be defined as a vector by taking $F_2 = 1$. DRAGON can be applied to univariate signals by taking $N=1$. We partition our time indexes into three groups of successive time steps and split accordingly $\mathcal{D}$ into three datasets: $\mathcal{D}_{train}$, $\mathcal{D}_{valid}$ and $\mathcal{D}_{test}$.

After choosing an architecture $\alpha$ and a set of hyperparamaters $\lambda$, we build the DNN $f^{\alpha, \lambda}$ and use $\mathcal{D}_{train}$ to train $f^{\alpha, \lambda}$ and optimize its weights $\theta$ by stochastic gradient descent:

\begin{center}
    $\hat{\theta} \in \underset{\theta \in \Theta(\alpha, \lambda)}{\argmin}\big(\ell(f_{\theta}^{\alpha, \lambda}, \mathcal{D}_{train})\big)$.
\end{center}

The forecast error of the DNN parameterized by $\hat{\theta}$ on $\mathcal{D}_{valid}$ is used to assess the performance of the selected $\alpha$ and $\lambda$. The best architecture and hyperparameters are optimized by solving:

\begin{center}
    $(\hat{\alpha}, \hat{\lambda}) \in \underset{\alpha \in \mathcal{A}}\argmin\Big(\underset{\lambda \in \Lambda(\alpha)}\argmin\big(\ell(f_{\hat{\theta}}^{\alpha, \lambda},\mathcal{D}_{valid})\big)\Big)$.
\end{center}

 The function $(\alpha, \lambda) \mapsto \ell(f_{\hat{\theta}}^{\alpha, \lambda}, \mathcal{D}_{valid})$ corresponds to the objective function of DRAGON. We finally will evaluate the performance of DRAGON by computing the forecast error on $\mathcal{D}_{test}$ using the DNN with the best architecture, hyperparameters and weights:

\begin{center}
    $\ell(f_{\hat{\theta}}^{\hat{\alpha}, \hat{\lambda}},\mathcal{D}_{test})$.
\end{center}

In practice, the second equation optimizing $\alpha$ and $\lambda$ can be solved separately or jointly. If we fix $\lambda$ for each $\alpha$, the optimization is made only on the architecture and is referred to as Neural Architecture Search (NAS). If $\alpha$ is fixed, then the optimization is only made on the model hyperparameters and is referred to as HyperParameters Optimization (HPO). DRAGON allows to fix $\alpha$ or $\lambda$ during parts of the optimization to perform a hierarchical optimization: ordering optimisation sequences during which only the architecture is optimised, and others during which only the hyperparameters are optimised.  In the following, we will describe our search space $\Omega = (\mathcal{A} \times \{\Lambda(\alpha), \alpha \in \mathcal{A}\})$.

\subsection{Architecture Search Space} \label{part sp archi}

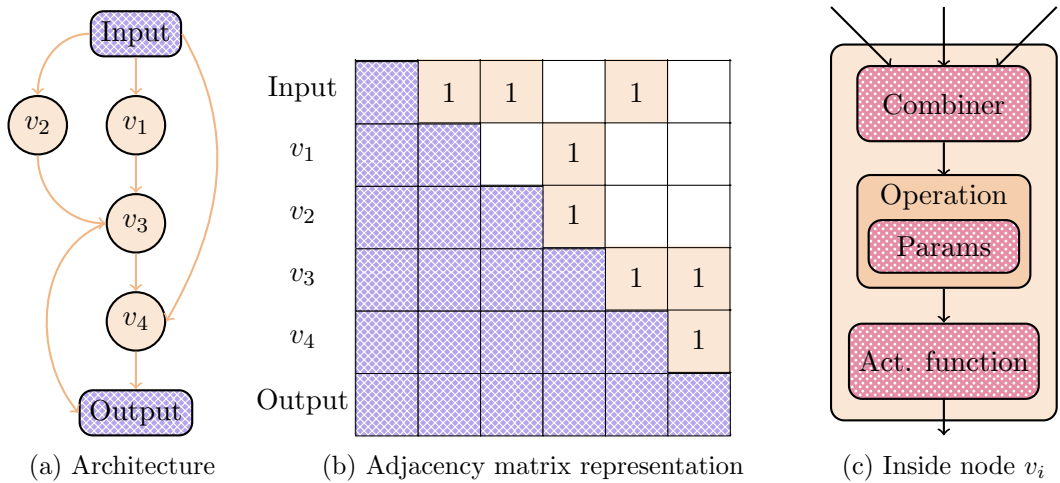
\begin{figure}[htbp]
    \centering
    \begin{subfigure}[b]{0.2\textwidth}
        \centering
        \begin{tikzpicture}[auto, thick,auto,main node/.style={circle,draw}]
            \draw
            node [in_out, preaction={fill, input_purple!40}, pattern color=white, pattern=crosshatch] (c1) {Input}
            node [archi_layer, fill=middle_beige, below=0.5cm of c1] (c2) {$v_1$}
            node [archi_layer,fill=middle_beige, left =0.5cm of c2] (c3) {$v_2$}
            node [archi_layer, fill=middle_beige, below=0.5cm of c2] (c4) {$v_3$}
            node [archi_layer, fill=middle_beige, below=0.5cm of c4] (c5) {$v_4$}
            node [in_out, preaction={fill, input_purple!40}, pattern color=white, pattern=crosshatch, below =0.5cm of c5] (out) {Output};
             \draw[->, middle_beige!300] (c1.west) to [out=180,in=90] (c3);
             \draw[->, middle_beige!300] (c1.east) to [out=300,in=60] (c5.east);
             \draw[->, middle_beige!300] (c1) -- (c2);
             \draw[->, middle_beige!300] (c2) -- (c4);
             \draw[->, middle_beige!300] (c4) -- (c5);
             \draw[->, middle_beige!300] (c3.south) to [out=270,in=180] (c4.west);
             \draw[->, middle_beige!300] (c5) -- (out);
             \draw[->, middle_beige!300] (c4.west) to [out=200,in=130] (out.west);
        \end{tikzpicture}
        \subcaption{Architecture}
        \label{fig:cell_gen_representation}
    \end{subfigure}
    \begin{subfigure}[b]{0.5\textwidth}
        \begin{tikzpicture}[box/.style={minimum size=0.83cm,draw}]
            \draw[fill=middle_beige, middle_beige] (0.83, 4.12) rectangle (1.62, 4.99);
            \draw[fill=middle_beige, middle_beige] (1.62, 4.99) rectangle (2.49, 4.16);
            \draw[fill=middle_beige, middle_beige] (3.32, 4.16) rectangle (4.16, 4.99);
            \draw[fill=middle_beige, middle_beige] (2.49, 4.16) rectangle (3.32, 3.32);
            \draw[fill=middle_beige, middle_beige] (2.49, 3.32) rectangle (3.32, 2.51);
            \draw[fill=middle_beige, middle_beige] (3.32, 2.51) rectangle (4.16, 1.62);
            \draw[fill=middle_beige, middle_beige] (4.16, 2.51) rectangle (4.99, 1.62);
            \draw[fill=middle_beige, middle_beige] (4.16, 1.62) rectangle (4.99, 0.83);
            \draw[step=0.83cm] (0,0) grid (5,5);
            \node at (-0.7, 4.6) {Input};
            \node at (-0.7, 3.76) {$v_1$};
            \node at (-0.7, 2.93) {$v_2$};
            \node at (-0.7, 2.1) {$v_3$};
            \node at (-0.7, 1.27) {$v_4$};
            \node at (-0.7, 0.44) {Output};
            \node at (1.245, 4.55) {1};
            \node at (2.075, 4.55) {1};
            \node at (3.735, 4.55) {1};
            \node at (2.905, 3.755) {1};
            \node at (2.905, 2.925) {1};
            \node at (3.735, 2.095) {1};
            \node at (4.565, 2.095) {1};
            \node at (4.565, 1.265) {1};
            \foreach \y in {1,...,6}
            \foreach \x in {1, ..., \y}
              {
                \node [box,preaction={fill, input_purple!40}, pattern color=white, pattern=crosshatch] at (0.83*\x-0.41, 6-0.83*\y-0.62){ };
            }
        \end{tikzpicture}
        \caption{Adjacency matrix representation}
        \label{fig:adj_matrix_representation}
    \end{subfigure}
    \begin{subfigure}[b]{0.2\textwidth}
        \centering
        \begin{tikzpicture}[auto, thick]
            \draw
            node [in_out, minimum height = 5cm, minimum width=3cm, fill=middle_beige] (global) {}
            node [in_out, minimum height = 1cm, minimum width=2.3cm, preaction={fill, output_red!50}, pattern color=white, pattern=crosshatch dots] at ([yshift=-0.8cm]global.north) {Combiner}
            node [in_out, minimum height = 1.5cm, minimum width=2.3cm, label={[anchor=north]north:Operation}, fill=middle_beige!200] at ([yshift=-2.5cm]global.north) {}
            node [in_out, minimum height = 0.7cm,minimum width=2cm,preaction={fill, output_red!50}, pattern color=white, pattern=crosshatch dots] at ([yshift=-2.7cm]global.north) {Params}
            node [in_out, minimum height = 1cm,minimum width=2.3cm, preaction={fill, output_red!50}, pattern color=white, pattern=crosshatch dots] at ([yshift=0.8cm]global.south) {Act. function};
            \draw[->](-1.5, 3) -- (-0.7, 2.2); 
            \draw[->](0, 3) -- (0, 2.2);
            \draw[->](1.5, 3) -- (0.7, 2.2);
            \draw[->](0, 1.2) -- (0, 0.75);
            \draw[->](0, -0.75) -- (0, -1.2);
            \draw[->](0, -2.2) -- (0, -2.7);
        \end{tikzpicture}
        \subcaption{Inside node $v_i$}
        \label{fig:node_representation}
    \end{subfigure}
    \caption{DNN encoding as a directed acyclic graph (DAG). The elements in blue (crosshatch) are fixed by the framework, the architecture elements from $\alpha$ are displayed in beige and the hyperparameters $\lambda$ are in pink (dots).}
\end{figure}

First, we define our architecture search space $\mathcal{A}$. We propose to model a DNN by a Directed Acyclic Graph (DAG) with a single input and output \citep{fiore2013algebra}. A DAG $\Gamma = (\mathcal{V}, \mathcal{E})$ is defined by its nodes (or vertices) set $\mathcal{V} = \{v_1, ..., v_n\}$ and its edges set $\mathcal{E} \subseteq \{(v_i, v_j) | v_i, v_j \in \mathcal{V} \}$. Each node $v$ represents a DNN layer as defined in Subsection~\ref{part dl related}, such as a convolution, a recurrence, or a matrix product. To eliminate isolated nodes, we impose each node to be connected by a path to the input and the output. The graph acyclicity implies a partial ordering of the nodes. If a path exists from the node $v_a$ to a node $v_b$, then we can define a relation order between them: $v_a < v_b$. Acyclicity prevents the existence of a path from $v_b$ to $v_a$. However, this relation order is not total. When dealing with symmetric graphs where all nodes are not connected, several nodes' ordering may be valid for the same graph. For example in Figure \ref{fig:cell_gen_representation}, the orderings $v_1 > v_2$ and $v_2 > v_1$ are both valid.
 
Hence, a DAG $\Gamma$ is represented by a sorted list $\mathcal{L}$, such that $|\mathcal{L}| = m$, containing the graph nodes, and its adjacency matrix $M \in \{0,1\}^{m\times m}$ \citep{zhang2019d}. The matrix $M$ is built such that: $M(i,j) = 1 \Leftrightarrow (v_i, v_j) \in \mathcal{E}$. Because of the graph's acyclicity, the matrix is upper triangular with its diagonal filled with zeros. The input node has no incoming connection, and the output node has no outcoming connection, meaning $\sum_{i=1}^{m} M_{i,1} = 0$ and $\sum_{j=1}^{m} M_{m,j} = 0$.  Besides, the input is necessarily connected to the first node and the last node to the output for any graph, enforcing: $M_{1, 2} = 1$ and $M_{m-1, m} = 1$. As isolated nodes do not exist in the graph, we need at least a non-zero value on every row and column, except for the first column and last row. We can express this property as: $\forall i < m: \sum_{j=i+1}^{m} M_{i,j} > 0$ and $\forall j > 1: \sum_{i=j+1}^{m} M_{i,j} > 0$.  Finally, the ordering of the partial nodes does not allow a bijective encoding: several matrices $M$ may encode the same DAG.

To summarize, we have $\mathcal{A} = \{\Gamma = (\mathcal{V}, \mathcal{E}) = (\mathcal{L}, M)\}$. The graphs $\Gamma$ are parameterized by their size $m$ which is not equal for all graphs. As we will see in Section~\ref{part: search algo} the DNNs size may vary during the optimization.

\subsection{Hyperparameters Search Space} \label{part sp hp}

For any fixed architecture $\alpha \in \mathcal{A}$, let's define our hyperparameters search space induced by $\alpha: \Lambda(\alpha)$. As mentioned above, the DAG nodes represent the DNN hidden layers. A set of hyperparameters $\lambda$, also called a graph node, is composed of a combiner, a layer operation and an activation function (see Figure \ref{fig:node_representation}). Each layer operation is associated with a specific set of parameters, like output or hidden dimensions, convolution kernel size or dropout rate. We provide in Appendix \ref{anx:op_hp} a table with all available operations and their associated parameters. The hyperparameters search space $\Lambda(\alpha)$ is made of sets $\lambda$ composed with a combiner, the layer's parameters and the activation function.

First, we need a combiner as each node can receive an arbitrary number of input connections. The parents' latent representations should be combined before being fed to the operation. Taking inspiration from the Google Brain Team Evolved Transformer \citep{so2019evolved}, we propose three types of combiners: element-wise addition, element-wise multiplication and concatenation. The input vectors may have different channel numbers and the combiner needs to level them. This issue is rarely mentioned in the literature, where authors prefer to keep a fixed channel number \citep{liu2018darts}. In the general case, for element-wise combiners, the combiner output channel matches the maximum channel number of latent representation. We apply zero-padding on the smaller inputs. For the concatenation combiner, we consider the sum of the channel number of each input. Some operations, for instance, the pooling and the convolution operators, have kernels. Their calculation requires that the number of channels of the input vector is larger than this kernel. In these cases, we also perform zero-padding after the combiner to ensure that we have the minimum number of channels required. We use the node order to create our DNN. We build the nodes one at a time, following their order in $\mathcal{L}$. The first node is created using the data input shape as argument. After its creation we compute and gather its output shape. Then, for each following node, we compute the layer operation input shape according to the output shapes of the connected nodes and the combiner. After building the operation we compute its output shape for the next layers. Finally, as depicted in Figure~\ref{fig:node_representation}, each node ends with an activation function. The hyperparameters optimized for each node can be found Annex~\ref{anx:op_hp}.

To summarize, we define every node as the sequence of combiner $\rightarrow$ layer $\rightarrow$ activation function. In our search space $\Lambda(\alpha)$, the nodes are encoded by Python objects having as attributes the combiner name, the layer corresponding to the operation set with the hyperparameters encoded as a PyTorch Module followed by the activation function. The set $\mathcal{L}$ is a variable-length list containing each node.

\section{Search algorithm} \label{part 4}

The search space from DRAGON $\Omega = (\mathcal{A} \times \{\Lambda(\alpha), \alpha \in \mathcal{A}\})$ defined in the previous section is a mixed and variable space: it may contain integers, float, and categorical values, and the dimension of its elements, the DNNs, is not fixed. We need to design a search algorithm able to efficiently navigate through this search space. While several metaheuristics can solve mixed and variable-size optimization problems \citep{talbi2023}, we chose to start with an evolutionary algorithm. This metaheuristic was the most intuitive for us to  manipulate directed acyclic graphs. It has been used to optimize graphs in other fields, for example on logic circuits \citep{aguirre2003evolutionary}. In Section~\ref{part5} we compare our model to other simple metaheuristics: the Random Search and the Simulated Annealing, but the design of more complex metaheuristics using our search space and their comparison with the evolutionary algorithm are left to future work.

\subsection{Evolutionary algorithm design} \label{part: search algo}

\begin{figure}[htbp]
    \centering
    \begin{tikzpicture}[auto, thick]
         \draw
            node [input] (init) {}
            node [in_out, minimum height = 1cm, minimum width=4.5cm, below of=init, fill=output_red!10] (crit) {Is the final criterium $B$ reached?}
            node [in_out, minimum height = 1cm, minimum width=4.5cm, below =1cm of crit, fill=output_red!20] (select) {Tournament selection}
            node [in_out, minimum height = 1cm, minimum width=4.5cm, below =0.4cm of select, fill=input_purple!60] (repro) {Reproduction:\\crossover and mutation}
            node [in_out, minimum height = 1cm, minimum width=4.5cm, below =0.4cm of repro, fill=output_red!30] (replacement) {Offsprings evaluation}
            node [in_out, minimum height = 1cm, minimum width=4.5cm, below =0.4cm of replacement, fill=output_red!40] (end) {Replacement}
            node[input, right =1.5cm of crit] (out){};

        \draw[->](init) node[above] {\raisebox{1ex}{\epsdice{3}} \raisebox{-1ex}{\rotatebox[origin=c]{45}{\epsdice{2}}} $\rightarrow$ Create and initialize a population of size $K$} -- (crit);
        \draw[->](crit) -- node[left]{No} (select);
        \draw[->](select) -- (repro);
        \draw[->](repro) -- (replacement);
        \draw[->](replacement) -- (end);
        \draw[->] (crit.east) -- node[below] {Yes} (out) node[right] {Best model};
        \draw[->] (end.west) -| (crit.191);
    \end{tikzpicture}
    \caption{Evolutionary algorithm flowchart.}
    \label{fig:ga_framework}
\end{figure}
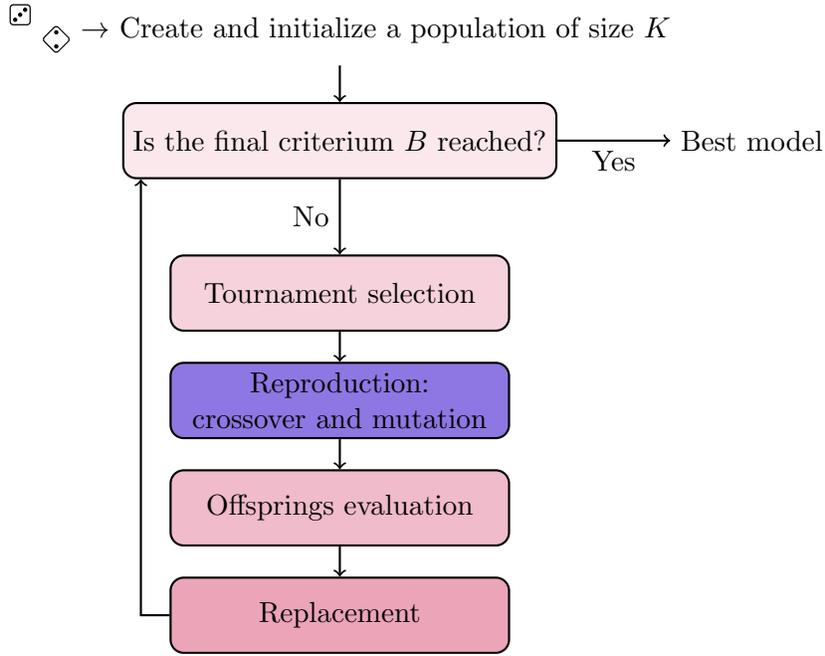

Evolutionary algorithms represent popular metaheuristics which are well adapted to solve mixed and variable-space optimization algorithms \citep{talbi2023}. They have been widely used for the automatic design of DNNs \citep{li2022survey}. The idea is to evolve a randomly generated population of DAGs to converge towards an optimal DNN. An optimal solution for a time series forecasting task is defined as a DNN minimizing a forecasting error. As training a DNN is expensive in time and computational resources, we implemented an asynchronous version, also called steady-state, of the evolutionary algorithm. This version is more efficient on High-Performance Computing (HPC) systems as detailed by \citet{liu2018hierarchical}. At the beginning of the algorithm, a set of $K$ random DNNs is generated. Each solution is train on $\mathcal{D}_{train}$ and evaluate on $\mathcal{D}_{valid}$ to create a population of size $K$. Then, for a certain number of iterations or a fixed time budget $B$, once a processus is free, a selection operator selects two solutions from the population. Those solutions are modified using crossover and mutation operators to create two offsprings. Those are trained and evaluated by the free process. Then, for each offspring, if its loss $\ell$ is less than the worst loss from the population, the offspring replaces the worst individual. Using an asynchronous version instead of the classical one avoids waiting for a whole generation to be evaluated and saves some time. The complete flowchart is shown in Figure \ref{fig:ga_framework}.

DRAGON's search space defined Section~\ref{part 3} is not directly efficient with common mutation and crossover operators. Therefore, we had to define evolution operators specific for our search space. Those operators can be used with various metaheuristics: a mutation operator for example can be used as a neighborhood operator for a local search. We split the operators into two categories: hyperparameters specific operators and architecture operators. The idea is to allow a sequential or joint optimization of the hyperparameters and the architecture. All the candidate operations which can be used in the the graphs nodes do not share the same hyperparameters. Thus, drawing a new layer means modifying all its parameters and one can lose the optimization made on the hyperparameters of the previous operation. Using sequential optimization, the algorithm can first find well-performing architectures and operations during the architecture search and then fine-tune the found DNNs during the hyperparameters search.

\subsection{Architecture evolution}

 In this section, we introduce the architecture-specific search operators from DRAGON. By architecture, we mean the search space $\mathcal{A}$ defined above: the nodes' operation and the edges between them. 
 
 \paragraph{Mutation.}The mutation operators are simple modifications inspired by the Graph Edit Distance \citep{abu2015exact}: insertion, deletion and substitution of both nodes and edges. Given a graph $\Gamma = (\mathcal{L}, M)$, the mutation operator will draw the set $\mathcal{L}' \subseteq \mathcal{L}$ and apply a transformation to each node of $\mathcal{L}'$. Let's have $v_i \in \mathcal{L}'$ the node that will be transformed:
\begin{itemize}
\item \textbf{Node insertion:} we draw a new node with its combiner, operation and activation function. We insert the new node in our graph at the position $i + 1$. We draw its incoming and outgoing edges by verifying that we do not generate an isolated node.
\item \textbf{Node deletion:} we delete the node $v_i$. In the case where it generates other isolated nodes, we draw new edges.
\item \textbf{Parents modification:} we modify the incoming edges for $v_i$ and make sure we always have at least one.
\item \textbf{Children modification:} we modify the outgoing edges for $v_i$ and make sure we always have at least one.
\item \textbf{Node modification:} we draw the new content of $v_i$, the new combiner, the operation and/or the activation function.
\end{itemize}
Thanks to these mutation operators, we make our search space connected, as explained in Section~\ref{part 2}. In fact, by successively using these operations, we can move from any graph to any other, since the Graph Edit Distance can be used with any pair of graphs.

\paragraph{Crossover.} The second architecture-specific operator we implemented is the crossover. The idea is to inherit patterns from two parents to create two offsprings. The original crossover is applied to two vectors. It exchanges two parts of these vectors. In our case, the individuals are graphs. Let's say we have two parents $\Gamma_1$ and $\Gamma_2$. The first step is to randomly select one subgraph from each parent, $\gamma_1 \subset \Gamma_1$ and $\gamma_2 \subset \Gamma_2$ to exchange (see figure \ref{fig:crossover_1}). Next the two offspring $\Gamma_1'$ and $\Gamma_2'$ are generated from $\Gamma_1$ and $\Gamma_2$ by removing $\gamma_1$ and $\gamma_2$, as shown in figure \ref{fig:crossover_2}. Next, we need to define the position at which each of the subgraphs will be inserted into the host graph. The idea is to preserve the overall structure of the graph. In other words, if the subgraph was at the beginning of the parent graph, it should also be at the beginning of the child graph, and vice versa. 
We denote here, for a node $v \in \Gamma$, $p(v, \Gamma)$ its position in the graph $\Gamma$, and $P(\Gamma) = \{p(v, \Gamma), v \in \Gamma\}$ the set of all nodes positions in $\Gamma$. We compute the future positions of each node $v \in \gamma_1$ in $\Gamma_2'$ sequentially, starting with the first node, $v_1 \in \argmin_{v \in \gamma_1} p(v, \gamma_1)$. The position $p(v_1, \Gamma_2')$ of $v_1$ in the graph $\Gamma_2'$ can be computed as:
\begin{align*}
    p(v_1, \Gamma_2') \in \underset{p \in P(\Gamma_2')}\argmin(|p - p(v_1, \Gamma_1)|)
\end{align*}
The positions from the following nodes $\{v_2, ..., v_g\} \in \gamma_1$ are computed to respect the structure of $\Gamma_1$ and $\gamma_1$:

\begin{align*}
    p{(v_i, \Gamma_2')} = \min\big(p{(v_i, \Gamma_1)} - p{(v_{i-1}, \Gamma_1)} + p{(v_{i-1}, \Gamma_2')}, |\Gamma_2'| + |\gamma_1|\big)
\end{align*}

\begin{figure}[htbp]
    \centering
    \begin{subfigure}[b]{\textwidth}
         \begin{tikzpicture}[auto, thick,auto,main node/.style={circle,draw}, box/.style={minimum size=0.5cm,draw}]
            \footnotesize
             \draw
             node [num_circle, fill=white] (num1) {1}
             
             node [in_out, above right=0.7cm and 1.5cm of num1, preaction={fill, input_purple!40}, pattern color=white, pattern=crosshatch] (in1) {Input}
             node [num_circle, below=0.35cm of in1] (a2) {$a_2$}
             node [num_circle, dotted, left=0.3cm of a2, fill=input_purple!70] (a1) {$a_1$}
             node [num_circle, right=0.3cm of a2] (a3) {$a_3$}
             node [in_out, below=0.35cm of a2, preaction={fill, input_purple!40}, pattern color=white, pattern=crosshatch] (ou1) {Output}
             node[in_out, below right=0.5cm and 0.5cm of ou1.center, draw=white, fill=white] (leg1) {Parent 1: $\Gamma_1$}
             
             node [in_out, above right=0.7cm and 9cm of num1, preaction={fill, output_red!40}, pattern color=white, pattern=crosshatch] (in2) {Input}
             node [a_dia, dotted, below=0.2cm of in2, fill=output_red!70] (b3) {$b_3$}
             node [a_dia, left=0.2cm of b3] (b1) {$b_1$}
             node [a_dia, right=0.2cm of b3] (b4) {$b_4$}
             node [a_dia, dotted, right = 0.7cm of in2.north, fill=output_red!70] (b2) {$b_2$}
             node [in_out, below=0.2cm of b3, preaction={fill, output_red!40}, pattern color=white, pattern=crosshatch] (ou2) {Output}
             node[in_out, below right=0.5cm and 0.7cm of ou2.center, draw=white, fill=white] (leg2) {Parent 2: $\Gamma_2$};
             \draw[->] (in1.west) to [out=180,in=90] (a1);
             \draw[->] (in1) -- (a2);
             \draw[->] (in1.east) to [out=0,in=90] (a3);
             \draw[->] (a1) to [out=270,in=160] (ou1.west);
             \draw[->] (a1) -- (a2);
             \draw[->] (a2) -- (a3);
             \draw[->] (a2) -- (ou1);
             \draw[->] (a3) to [out=270,in=20] (ou1.east);
    
            \draw[->] (in2.west) to [out=180,in=90] (b1);
            \draw[->] (b1) to [out=270,in=160] (ou2.west);
            \draw[->] (b3) -- (ou2);
            \draw[->] (b4) to [out=270,in=20] (ou2.east);
            \draw[->] (in2.north) to [out=90, in=180] (b2.north west);
            \draw[->] (b2) -- (b4);
            \draw[->] (b2.south) to [out=240, in=60] (b3.north east);
        
            \draw[step=0.5cm, color=gray!20] (4.5-0.001, 1.5) grid (7, -1);
            \foreach \y in {1,...,5}
                \foreach \x in {1, ..., \y}
                  {
                    \node [box, preaction={fill, input_purple!40}, pattern color=white, pattern=crosshatch, draw=input_purple!40] at (4.25+0.5*\x, 1.75-0.5*\y){ };
                }
            \draw[] (4.5-0.001, 1.5) rectangle (7, -1);
            \draw[dotted, line width=0.3mm, fill=input_purple!70] (5, 1.5) rectangle (5.5, -0.5);
            \draw[dotted, line width=0.3mm, fill=input_purple!70] (4.5, 1) rectangle (6.5, 0.5);
            \node at (5.25, 1.25) {1};
            \node at (5.25, 0.75) {0};
            \node at (5.25, 0.25) {0};
            \node at (5.25, -0.25) {0};
            \node at (4.75, 0.75) {0};
            \node at (5.75, 0.75) {1};
            \node at (6.25, 0.75) {0};
            \node at (6.75, 0.75) {1};
            \node at (5.75, 1.25) {1};
            \node at (6.25, 1.25) {1};
            \node at (6.75, 1.25) {0};
            \node at (6.25, 0.25) {1};
            \node at (6.75, 0.25) {1};
            \node at (6.75, -0.25) {1};
    
            \draw[step=0.5cm, color=gray!20] (12.5-0.001, 2) grid (15.5, -1);
            \foreach \y in {1,...,6}
                \foreach \x in {1, ..., \y}
                  {
                    \node [box, preaction={fill, output_red!40}, pattern color=white, pattern=crosshatch, draw=output_red!40] at (12.25+0.5*\x, 2.25-0.5*\y){ };
                }
            \draw[] (12.5-0.001, 2) rectangle (15.5, -1);
            \draw[dotted, line width=0.3mm, fill=output_red!70] (13.5, 1.5) rectangle (14, -1);
            \draw[dotted, line width=0.3mm, fill=output_red!70] (14, 1.5) rectangle (14.5, -1);
            \draw[dotted, line width=0.3mm, fill=output_red!70] (13, 1) rectangle (15.5, 0.5);
            \draw[dotted, line width=0.3mm, fill=output_red!70] (13, 0.5) rectangle (15.5, 0);
            \node at (13.25, 1.75) {1};
            \node at (13.75, 1.75) {1};
            \node at (14.25, 1.75) {0};
            \node at (14.75, 1.75) {0};
            \node at (15.25, 1.75) {0};
            \node at (13.75, 1.25) {0};
            \node at (14.25, 1.25) {0};
            \node at (14.75, 1.25) {0};
            \node at (15.25, 1.25) {1};
            \node at (13.25, 0.75) {0};
            \node at (13.75, 0.75) {0};
            \node at (14.25, 0.75) {1};
            \node at (14.75, 0.75) {1};
            \node at (15.25, 0.75) {0};
            \node at (13.25, 0.25) {0};
            \node at (13.75, 0.25) {0};
            \node at (14.25, 0.25) {0};
            \node at (14.75, 0.25) {0};
            \node at (15.25, 0.25) {1};
            \node at (13.75, -0.25) {0};
            \node at (14.25, -0.25) {0};
            \node at (15.25, -0.25) {1};
            \node at (13.75, -0.75) {0};
            \node at (14.25, -0.75) {0};
            
        \end{tikzpicture}

    \subcaption{\centering 1st step: we select the two subgraphs $\gamma_1 \subset \Gamma_1$ and $\gamma_2 \subset \Gamma_2$ that would be exchanged. They are highlighted with dotted lines and darker colors.}
    \label{fig:crossover_1}
    \end{subfigure}
     \begin{subfigure}[b]{\textwidth}
         \begin{tikzpicture}[auto, thick,auto,main node/.style={circle,draw}, box/.style={minimum size=0.5cm,draw}]
            \footnotesize
             \draw
             node [num_circle, fill=white] (num1) {2}
             
             node [in_out, above right=0.7cm and 1.5cm of num1, preaction={fill, input_purple!40}, pattern color=white, pattern=crosshatch] (in1) {Input}
             node [num_circle, below=0.35cm of in1] (a2) {$a_2$}
             node [num_circle, right=0.3cm of a2] (a3) {$a_3$}
             node [in_out, below=0.35cm of a2, preaction={fill, input_purple!40}, pattern color=white, pattern=crosshatch] (ou1) {Output}
             node[in_out, below right=0.5cm and 0.5cm of ou1.center, draw=white, fill=white] (leg1) {Offspring 1: $\Gamma_1'$}
             
             node [in_out, above right=0.7cm and 9cm of num1, preaction={fill, output_red!40}, pattern color=white, pattern=crosshatch] (in2) {Input}
             node [a_dia, dotted, below=0.2cm of in2, white] (b3) {b3}
             node [a_dia, left=0.2cm of b3] (b1) {$b_1$}
             node [a_dia, right=0.2cm of b3] (b4) {$b_4$}
             node [in_out, below=0.2cm of b3, preaction={fill, output_red!40}, pattern color=white, pattern=crosshatch] (ou2) {Output}
             node[in_out, below right=0.5cm and 0.7cm of ou2.center, draw=white, fill=white] (leg2) {Offspring 2: $\Gamma_2'$};
             \draw[->] (in1) -- (a2);
             \draw[->] (in1.east) to [out=0,in=90] (a3);
             \draw[->] (a2) -- (a3);
             \draw[->] (a2) -- (ou1);
             \draw[->] (a3) to [out=270,in=20] (ou1.east);
    
            \draw[->] (in2.west) to [out=180,in=90] (b1);
            \draw[->] (b1) to [out=270,in=160] (ou2.west);
            \draw[->] (b4) to [out=270,in=20] (ou2.east);
        
            \draw[step=0.5cm, color=gray!20] (4.5-0.001, 1.5) grid (6.5, -0.5);
            \foreach \y in {1,...,4}
                \foreach \x in {1, ..., \y}
                  {
                    \node [box, preaction={fill, input_purple!40}, pattern color=white, pattern=crosshatch, draw=input_purple!40] at (4.25+0.5*\x, 1.75-0.5*\y){ };
                }
            \draw[] (4.5-0.001, 1.5) rectangle (6.5, -0.5);
            \node at (5.25, 1.25) {1};
            \node at (5.75, 0.75) {1};
            \node at (6.25, 0.75) {1};
            \node at (5.75, 1.25) {1};
            \node at (6.25, 1.25) {0};
            \node at (6.25, 0.25) {1};
    
            \draw[step=0.5cm, color=gray!20] (12.5-0.001, 1.5) grid (14.5, -0.5);
            \foreach \y in {1,...,4}
                \foreach \x in {1, ..., \y}
                  {
                    \node [box,box, preaction={fill, output_red!40}, pattern color=white, pattern=crosshatch, draw=output_red!40] at (12.25+0.5*\x, 1.75-0.5*\y){ };
                }
            \draw[] (12.5-0.001, 1.5) rectangle (14.5, -0.5);
            \node at (13.25, 1.25) {1};
            \node at (13.75, 1.25) {0};
            \node at (14.25, 1.25) {0};
            \node at (13.75, 0.75) {0};
            \node at (14.25, 0.75) {1};
            \node at (14.25, 0.25) {1};
            
        \end{tikzpicture}
    \subcaption{\centering 2nd step: create the two offsprings $\Gamma_1'$ and $\Gamma_2'$ from $\Gamma_1$ and $\Gamma_2$ by removing $\gamma_1$ and $\gamma_2$.}
    \label{fig:crossover_2}

    \end{subfigure}
     \begin{subfigure}[b]{\textwidth}
         \begin{tikzpicture}[auto, thick,auto,main node/.style={circle,draw}, box/.style={minimum size=0.5cm,draw}]
            \footnotesize
             \draw
             node [num_circle, fill=white] (num1) {3}
             
             node [in_out, above right=0.7cm and 1.5cm of num1, preaction={fill, input_purple!40}, pattern color=white, pattern=crosshatch] (in1) {Input}
             node [num_circle, below=0.35cm of in1] (a2) {$a_2$}
             node [a_dia, dotted, left = 0.7cm of in1.north, fill=output_red!70] (b2) {$b_2$}
             node [a_dia, dotted, left=0.35cm of a2, fill=output_red!70] (b3) {$b_3$}
             node [num_circle, right=0.3cm of a2] (a3) {$a_3$}
             node [in_out, below=0.35cm of a2, preaction={fill, input_purple!40}, pattern color=white, pattern=crosshatch] (ou1) {Output}
             node[in_out, below right=0.5cm and 0.5cm of ou1.center, draw=white, fill=white] (leg1) {Offspring 1: $\Gamma_1'$}
             
             node [in_out, above right=0.7cm and 9cm of num1, preaction={fill, output_red!40}, pattern color=white, pattern=crosshatch] (in2) {Input}
            node [num_circle, dotted, below=0.3cm of in2, fill=input_purple!70] (a1) {$a_1$}
             node [a_dia, left=0.3cm of a1] (b1) {$b_1$}
             node [a_dia, right=0.3cm of a1] (b4) {$b_4$}
             node [in_out, below=0.3cm of a1, preaction={fill, output_red!40}, pattern color=white, pattern=crosshatch] (ou2) {Output}
             node[in_out, below right=0.5cm and 0.7cm of ou2.center, draw=white, fill=white] (leg2) {Offspring 2: $\Gamma_2'$};
             \draw[->] (in1) -- (a2);
             \draw[->] (in1.east) to [out=0,in=90] (a3);
             \draw[->] (a2) -- (a3);
             \draw[->] (a2) -- (ou1);
             \draw[->] (a3) to [out=270,in=20] (ou1.east);
             \draw[->, dotted] (in1.north) to [out=90, in=0] (b2.north east);
             \draw[->, dotted] (b2) -- (b3);
             \draw[->, dotted] (b2.south) to [out=300, in=160] (a2.north west);
             \draw[->, dotted] (b3.south east) to [out=330, in=210] (a3.south west);

            \draw[->] (in2.west) to [out=180,in=90] (b1);
            \draw[->] (b1) to [out=270,in=160] (ou2.west);
            \draw[->] (b4) to [out=270,in=20] (ou2.east);
            \draw[->, dotted] (b1) -- (a1);
            \draw[->, dotted] (a1) -- (b4);
        
            \draw[step=0.5cm, color=gray!20] (4.5-0.001, 2) grid (7.5, -1);
            \foreach \y in {1,...,6}
                \foreach \x in {1, ..., \y}
                  {
                    \node [box, preaction={fill, input_purple!40}, pattern color=white, pattern=crosshatch, draw=input_purple!40] at (4.25+0.5*\x, 2.25-0.5*\y){ };
                }
            \draw[] (4.5-0.001, 2) rectangle (7.5, -1);
            \draw[dotted, line width=0.3mm, fill=output_red!70] (5, 2) rectangle (5.5, -0.5);
            \draw[dotted, line width=0.3mm, fill=output_red!70] (5.5, 2) rectangle (6, -0.5);
            \draw[dotted, line width=0.3mm, fill=output_red!70] (4.5, 1.5) rectangle (7, 1);
            \draw[dotted, line width=0.3mm, fill=output_red!70] (4.5, 1) rectangle (7, 0.5);
            \node at (5.25, 1.75) {\textbf{\textcolor{white}{1}}};
            \node at (5.75, 1.75) {0};
            \node at (6.25, 1.75) {1};
            \node at (6.75, 1.75) {1};
            \node at (7.25, 1.75) {0};

            \node at (5.25, 1.25) {0};
            \node at (5.25, 0.75) {0};
            \node at (5.25, 0.25) {0};
            \node at (5.25, -0.25) {0};
            \node at (4.75, 0.75) {0};
            \node at (5.75, 0.75) {0};
            \node at (6.25, 0.75) {0};
            \node at (6.75, 0.75) {1};
            \node at (7.25, 0.75) {\textbf{\textcolor{output_red!70}{0}}};
            \node at (7.25, 1.25) {\textbf{\textcolor{output_red!70}{0}}};
            \node at (5.75, 1.25) {1};
            \node at (6.25, 1.25) {1};
            \node at (6.75, 1.25) {0};
            \node at (6.75, 0.25) {1};
            \node at (7.25, 0.25) {1};
            \node at (7.25, -0.25) {1};
            \node at (5.75, 0.25) {0};
            \node at (5.75, -0.25) {0};
            \node at (4.75, 1.25) {0};
            \node at (8.2, 1.5) {\large{\raisebox{1ex}{\epsdice{3}} \raisebox{-1ex}{\rotatebox[origin=c]{45}{\epsdice{2}}}}};
            \draw[->, color=output_red!70] (8.1, 1.4) to [out=270,in=0] (7.5, 1.25);
            \draw[->, color=output_red!70] (8.1, 1.4) to [out=270,in=0] (7.5, 0.75);

            \draw[step=0.5cm, color=gray!20] (12.5-0.001, 1.5) grid (15, -1);
            \foreach \y in {1,...,5}
                \foreach \x in {1, ..., \y}
                  {
                    \node [box, preaction={fill, output_red!40}, pattern color=white, pattern=crosshatch, draw=output_red!40] at (12.25+0.5*\x, 1.75-0.5*\y){ };
                }
            \draw[] (12.5-0.001, 1.5) rectangle (15, -1);
            \draw[dotted, line width=0.3mm, fill=input_purple!70] (13.5, 1) rectangle (14, -1);
            \draw[dotted, line width=0.3mm,  fill=input_purple!70] (13, 0.5) rectangle (15, 0);
            \node at (13.25, 1.25) {1};
            \node at (13.75, 1.25) {\textcolor{input_purple!70}{\textbf{0}}};
            \node at (14.25, 1.25) {0};
            \node at (14.75, 1.25) {0};
            \node at (13.75, 0.75) {1};
            \node at (14.25, 0.75) {0};
            \node at (14.75, 0.75) {1};
            \node at (13.25, 0.25) {0};
            \node at (13.75, 0.25) {0};
            \node at (14.25, 0.25) {1};
            \node at (14.75, 0.25) {0};
            \node at (13.75, -0.25) {0};
            \node at (14.75, -0.25) {1};
            \node at (13.75, -0.75) {0};
            \node at (14.5, 2) {\large{\raisebox{1ex}{\epsdice{3}} \raisebox{-1ex}{\rotatebox[origin=c]{45}{\epsdice{2}}}}};
            \draw[->, color=input_purple!70] (14, 2) to [out=180,in=90] (13.75, 1.5);
        \end{tikzpicture}
    \subcaption{\centering 3rd step: we insert the nodes from $\gamma_1$ in $\Gamma_2'$ and the nodes from $\gamma_2$ in $\Gamma_1'$ and reconstruct the edges.}
    \label{fig:crossover_3}

    \end{subfigure}
\caption{Crossover operator illustration.}
\label{fig:crossover}
\end{figure}

Finally, as shown Figure \ref{fig:cell_gen_representation}, the rows and columns corresponding to the nodes from $\gamma_1$ and $\gamma_2$ are inserted in the adjacency matrices of $\Gamma_2'$ and $\Gamma_1'$ at the previously computed positions. If the process has generated orphan nodes, we randomly generate the necessary connections.

\subsection{Hyperparameters evolution}

One of the architecture mutations consists in disturbing the node content. In this case, the node content is modified, including the operation. A new set of hyperparameters is then drawn. To refine this search, we defined specific mutations for the search space $\Lambda(\alpha)$. In the hyperparameters case, edges and nodes number are not affected. As for architecture-specific mutation, the operator will draw the set $\mathcal{L}' \subseteq \mathcal{L}$ and apply a transformation on each node of $\mathcal{L}'$. For each node $v_i$ from $\mathcal{L}'$, we draw $h_i$ hyperparameters, which will be modified by a neighbouring value. The hyperparameters in our search space belong to three categories:

\begin{itemize}
    \item \textbf{Categorical values:} the new value is randomly drawn among the set of possibilities deprived of the actual value. For instance, the activation functions, combiners, and recurrence types (LSTM/GRU) belong to this type of categorical variable.
    \item \textbf{Integers:} we select the neighbours inside a discrete interval around the actual value. For instance, it has been applied to convolution kernel size and output dimension.
    \item \textbf{Float:} we select the neighbours inside a continuous interval around the actual value. Such a neighbourhood has been defined for instance to the dropout rate.
\end{itemize}

\section{Experimental study} \label{part5}

\subsection{Baseline}

We compared our framework to two baselines. The first consists of 15 handcrafted models \citep{godahewa2021monash}, the second is more recent and compares 6 AutoML frameworks specifically designed for time series forecasting \citep{shchur2023autogluon}.

\paragraph{Handcrafted models} These are statistical, machine learning and deep learning models that were built and optimised by hand. We first have 5 traditional univariate forecasting models: Simple Exponential Smoothing (SES), Exponential Smoothing (ETS), Theta, Trigonometric Box-Cox ARMA Trend Seasonal (TBATS), Dynamic Harmonic Regression ARIMA (DHR-ARIMA) and 8 global forecasting models: Pooled Regression (PR), CatBoost, Prophet, Feed-Forward Neural Network (FFNN), N-BEATS, WaveNet, Transformer and DeepAR. The last 5 models are Deep Neural Networks. Refer to the original paper \citep{godahewa2021monash} and the Monash Time Series Forecasting Repository website\footnote{\url{https://forecastingdata.org/}} for more information about the models and their implementation. Finally, \citet{shchur2023autogluon} also provides the univariate forecasting model SeasonalNaive and the global deep learning model Temporal Fusion Transformer (TFT).

\paragraph{AutoML frmaeworks} \citet{shchur2023autogluon} compares 6 AutoML frameworks specifically designed for time series forecasting. They first used 4 statistical based AutoML framworks: AutoARIMA, AutoETS and AutoTheta which automatically tune the hyperparameters of the ARIMA, ETS and Theta models respectively for each time series individually based on an information criterion and StatEnsemble which takes the the forecasts median of three statistical models. Then, they included the AutoDL framework AutoPyTorch-Forecasting, which builds DNNs using a combination of Bayesian and multi-fidelity optimization and then leverages model ensemble. Finally, AutoGluon-TS, the AutoML framework proposed by \citet{shchur2023autogluon} relies on ensembling techniques rather than HPO or NAS. The models ensemble are made of local models such as ARIMA, Theta, ETS and SeasonalNaive, as well as global models such as DeepAR, PatchTST and Temporal Fusion Transformer. While it is interesting to compare ourselves with these state-of-the-art AutoML techniques, it is worth remembering that our framework does not yet provide an ensembling technique, and the scores obtained during optimization are based on the predictions of a single DNN.

\subsection{Experimental protocol}

We evaluated DRAGON on the established benchmark of Monash Time Series Forecasting Repository \citep{godahewa2021monash}. This archive contains a benchmark of more than 40 datasets, from which we selected the 27 that \citet{shchur2023autogluon} used fot their experiments. The time series are of different kinds and have variable distributions. More information on each dataset from the archive is available \ref{part info monash}. This task diversity allows to test DRAGON generalization and robustness abilities.

For these experiments, we configured our algorithm to have a population of $K=100$ individuals and we set the total budget to $B=$ 8 hours. We investigated a joint optimization of the architecture $\alpha$ and the hyperparameters $\lambda$. We ran our experiments on 5 cluster nodes, each equipped with 4 Tesla V100 SXM2 32GB GPUs, using Pytorch 1.11.0 and Cuda 10.2.

We took the data, the data generation functions, the training parameters (batch size, number of epochs, learning rate), the training and prediction functions from the Monash Time Series Forecasting Repository, and we only changed the models themselves. We also kept for each time series the forecast horizon and the lag used in the repository. We believe our comparison is fair to the handcrafted and automatically designed models. Finally, to evaluate the models' performance, we used the same metric and metric implementation as in the repository. This metric represents the forecast error $\ell$, and is the Mean Absolute Scaled Error (MASE), an absolute mean error divided by the average difference between two consecutive time steps \citep{hyndman2006another}. Given a time series $Y = (\mathbf{y}_1, ..., \mathbf{y}_n)$ and the predictions $\hat{Y} = (\mathbf{\hat{y}}_1, ..., \mathbf{\hat{y}}_n)$, the MASE can be defined as:
\begin{center}
    $\mathrm{MASE}(Y, \hat{Y}) = \frac{n-1}{n} \times \frac{\sum_{t=1}^{n}|\mathbf{y}_t - \hat{\mathbf{y}}_t|}{\sum_{t=2}^{n}|\mathbf{y}_t - \mathbf{y}_{t-1}|}$.
\end{center}

In our case, for $f \in \Omega$, $\mathcal{D}_0 = (X_0, Y_0) \subseteq \mathcal{D}$, we have $\ell(Y_0, f(X_0) = \mathrm{MASE}\big(Y_0, f(X_0)\big)$.

\subsection{Search Space}

\begin{figure}[htbp]
    \centering
    \begin{tikzpicture}[auto, thick]
        \draw
        node [in_out, minimum height = 1cm, minimum width=4cm, fill=input_purple, text=white](input) {Time Series input}
        node [in_out, minimum height = 4cm, minimum width=4.5cm, below=0.5cm of input, fill=middle_beige!30, label={[align=center, anchor=north]north:DNN:$f^{\alpha, \lambda}_{\theta}$}] (dnn) {}
        node [in_out, minimum height = 1.5cm, minimum width=4cm , below=1.3cm of input] (dag) {Directed Acyclic\\ Graph: $ \Gamma = f^{\alpha, \lambda}$}
         node [in_out, minimum height = 1cm, minimum width=4cm, below=0.3cm of dag, fill=output_red, text=white] (mlp) {Multi-layer Perceptron}
         node [in_out, minimum height = 1cm, minimum width=4cm, below=0.5cm of dnn, fill=output_red!30] (out) {Time Series Prediction};
        \draw[->](input) -- (dnn); 
        \draw[->](dag) -- (mlp);
        \draw[->](dnn) -- (out);
    \end{tikzpicture}
    \caption{Meta-architecture for Monash time series datasets.}
    \label{fig:metamodel_monash}
\end{figure}
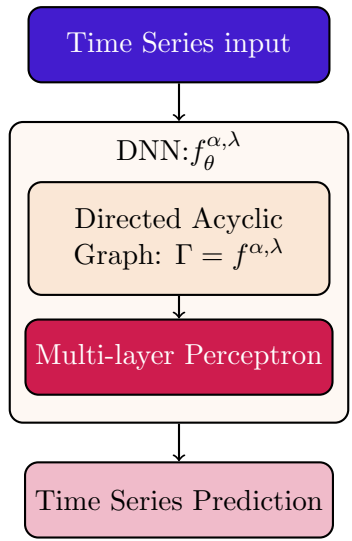

The generic search space defined Section~\ref{part 3} introduces a brick, the Directed Acyclic Graph, which cannot be directly defined as our search space. Instead, we defined a meta-architecture as represented Figure~\ref{fig:metamodel_monash}, which can be directly used to replace the repository's models. The graph $\Gamma$ can be composed with various one-dimensional candidate operations (e.g. 1D convolution, LSTM, MLP), which can be found with their associated hyperparameters Table~\ref{anx:op_hp}. Our DAG is followed by a Multi-layer Perceptron (MLP) which is used to retrieve the time series output dimension, as the number of channels may vary within $\Gamma$.
This search space is designed specifically for time series forecasting, but it could be modified for other tasks. For example if we want to use it for image classification, we would need a first graph with two-dimensional candidate operations, followed by a flatten layer, followed by a second graph with one-dimensional candidate operations and a final MLP layer.

\subsection{Results}

\begin{table}[htbp]
\begin{center}
\caption{Performance comparison of the baseline algorithms with DRAGON (based on the MASE metric) on 27 datasets. Wins corresponds to the number of datasets where the method produced a smaller loss than DRAGON, Losses corresponds to the number of datasets where the method produced a larger loss than DRAGON, Champion corresponds to the number of datasets where the method produced the smallest loss, and Failures corresponds to the number of datasets where the method failed.}
\small
\label{tab:results_summary}
\begin{tabular}{|c| c c c c |}
\hline
Algorithm(s) & Wins & Losses & Champion & Failures \\
\hline
SES & 1 & 25 & 0 & 1 \\
Theta & 6 & 20 & 0 & 1 \\
TBATS & 6 & 20 & 0 & 1 \\
ETS & 8 & 18 & 1 & 1 \\
(DHR-)ARIMA & 4 & 21 & 0 & 2 \\
PR & 1 & 25 & 0 & 1 \\
CatBoost & 1 & 25 & 0 & 1 \\
FFNN & 0 & 26 & 0 & 1 \\
N-BEATS & 0 & 26 & 0 & 1 \\
WaveNet & 2 & 23 & 0 & 2 \\
Transformer & 0 & 26 & 0 & 1 \\
DeepAR & 1 & 25 & 1 & 1 \\
TFT & 4 & 23 & 0 & 0 \\
SeasonalNaive & 1 & 26 & 0 & 0 \\
Prophet & 2 & 24 & 1 & 1 \\
AutoPytorch & 7 & 20 & 0 & 0 \\
AutoARIMA & 4 & 20 & 0 & 3 \\
AutoETS & 8 & 19 & 0 & 0 \\
AutoTheta & 9 & 16 & 0 & 2 \\
StatEnsemble & 9 & 15 & 3 & 3 \\
AutoGluon & 13 & 14 & 10 & 0 \\
DRAGON & - & - & 11 & 0 \\
\hline
\end{tabular}
\end{center}
\end{table}

We report a summary of the results in Table~\ref{tab:results_summary}. According to this summary, DRAGON outperforms all algorithms on 11 out of 27 datasets (41\%). No algorithm from the baseline was able to beat DRAGON on at least 50\% of the datasets and AutoGluon was the only algorithm able to beat DRAGON on more than a third of the datasets. The direct competitor of DRAGON, namely AutoPytorch which is another AutoDL framework, was only able to beat it on 7 datasets out of 27 (26\%). More detailed results can be found Table~\ref{tab:results_monash}.

\begin{table}[htbp]
\begin{center}
\caption{\footnotesize{Mean MASE for each dataset. We did not report all the individual scores from the handcrafted baseline, but the best score from the 15 models for each time series. The grayed values correspond to the minimal loss for the corresponding dataset.}}
\footnotesize
\label{tab:results_monash}
\begin{adjustbox}{angle=90}
\begin{tabular}{|c| c c c c c c c c|}
\hline
 Dataset & Handcrafted & AutoPytorch & AutoARIMA & AutoETS & AutoTheta & StatEnsemble & AutoGluon & DRAGON \\
\hline
COVID & 5.192 & 4.911 & 6.029 & 5.907 & 7.719 & 5.884 & 5.805 & \cellcolor{gray!20} 4.535 \\
Carparts & 0.746 & 0.746 & 1.118 & 1.133 & 1.208 & 1.052 & 0.747 & \cellcolor{gray!20} 0.745 \\
Electricity Hourly & 1.389 & 1.420 & - & 1.465 & - & - & \cellcolor{gray!20} 1.227 & 1.314 \\
Electricity Weekly & 0.769 & 2.322 & 3.009 & 3.076 & 3.113 & 3.077 & 1.892 & \cellcolor{gray!20} 0.644 \\
FRED-MD & \cellcolor{gray!20} 0.468 & 0.682 & 0.478 & 0.505 & 0.564 & 0.498 & 0.656 & 0.494 \\
Hospital & \cellcolor{gray!20} 0.673 & 0.770 & 0.820 & 0.766 & 0.764 & 0.753 & 0.741 & 0.750 \\
KDD & 0.844 & 0.764 & - & 0.988 & 1.010 & - & 0.709 & \cellcolor{gray!20} 0.678 \\
M1 Monthly & 1.074 & 1.278 & 1.152 & 1.083 & 1.092 & \cellcolor{gray!20} 1.045 & 1.235 & 1.069 \\
M1 Quarterly & 1.658 & 1.813 & 1.770 & 1.665 & 1.667 & 1.622 & \cellcolor{gray!20} 1.615 & 1.717 \\
M1 Yearly & 3.499 & 3.407 & 3.870 & 3.950 & 3.659 & 3.769 & \cellcolor{gray!20} 3.371 & 3.683 \\
M3 Monthly & 0.861 & 0.956 & 0.934 & 0.867 & 0.855 & 0.845 & \cellcolor{gray!20} 0.822 & 0.900 \\
M3 Other & 1.814 & 1.871 & 2.245 & 1.801 & 2.009 & \cellcolor{gray!20} 1.769 & 1.837 & 2.144 \\
M3 Quarterly & 1.117 & 1.180 & 1.419 & 1.121 & 1.119 & 1.096 & \cellcolor{gray!20} 1.057 & 1.087 \\
M3 Yearly & 2.774 & 2.691 & 3.159 & 2.695 & 2.608 & 2.627 & \cellcolor{gray!20} 2.520 & 2.775 \\
M4 Daily & 1.141 & 1.152 & 1.153 & 1.228 & 1.149 & 1.145 & 1.156 & \cellcolor{gray!20} 1.056 \\
M4 Hourly & 1.193 & 1.345 & 1.029 & 1.609 & 2.456 & 1.157 & \cellcolor{gray!20} 0.807 & 1.155 \\
M4 Monthly & 0.947 & 0.851 & 0.812 & 0.803 & 0.834 & \cellcolor{gray!20} 0.780 & 0.782 & 0.991 \\
M4 Quarterly & 1.161 & 1.176 & 1.276 & 1.167 & 1.183 & 1.148 & \cellcolor{gray!20} 1.139 & 1.190 \\
M4 Weekly & 0.453 & 2.369 & 2.355 & 2.548 & 2.608 & 2.375 & 2.035 & \cellcolor{gray!20} 0.446 \\
NN5 Daily & 0.789 & 0.807 & 0.935 & 0.870 & 0.878 & 0.859 & \cellcolor{gray!20} 0.761 & 0.892 \\
NN5 Weekly & 0.808 & 0.865 & 0.998 & 0.980 & 0.963 & 0.977 & 0.860 & \cellcolor{gray!20} 0.703 \\
Pedestrians & 0.247 & 0.354 & - & 0.553 & - & - & 0.312 & \cellcolor{gray!20} 0.218 \\
Tourism Monthly & \cellcolor{gray!20} 1.409 & 1.495 & 1.585 & 1.529 & 1.666 & 1.469 & 1.442 & 1.434 \\
Tourism Quarterly & 1.475 & 1.647 & 1.655 & 1.578 & 1.648 & 1.539 & 1.537 & \cellcolor{gray!20} 1.471 \\
Tourism Yearly & 2.590 & 3.004 & 4.044 & 3.183 & 2.992 & 3.231 & 2.946 & \cellcolor{gray!20} 2.337 \\
Vehicle Trips & 1.176 & 1.162 & 1.427 & 1.301 & 1.284 & 1.203 & \cellcolor{gray!20} 1.113 & 1.645 \\
Web Traffic & 0.973 & 0.962 & 1.189 & 1.207 & 1.108 & 1.068 & 0.938 & \cellcolor{gray!20} 0.561 \\
\hline
\end{tabular}
\end{adjustbox}
\end{center}
\end{table}

\begin{figure}[htbp]
    \centering
    \includegraphics[width=0.9\textwidth]{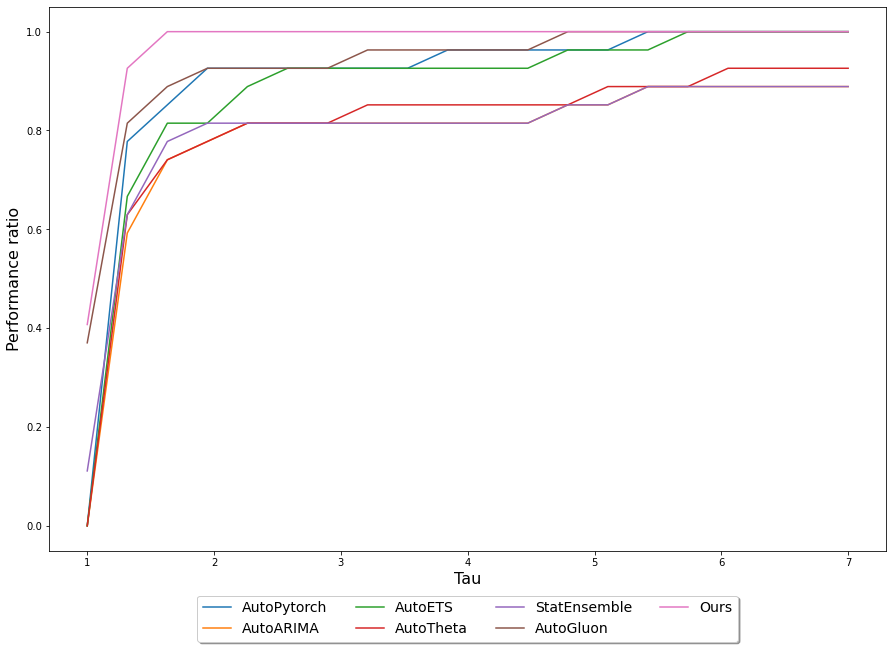}
    \caption{Performance profile $\rho_{s}(\tau)$ for each algorithm $s$ from the AutoML baseline, with $\tau \in [1, 7]$.}
    \label{fig:perf_ratio}
\end{figure}

To have a more visual comparison of the different algorithms from the baseline, we used the performance profile as defined by \citet{dolan2002benchmarking}. We name $\mathcal{P}$ the set of the 27 datasets, $\mathcal{S}$ the set of the 22 algorithms from the baseline and $l_{p,s}$ the final score (loss) of the algorithm $s \in \mathcal{S}$ on the dataset $p \in \mathcal{P}$. We define the performance ratio $r_{p,s}$ of $s$ on $p$ as:
\begin{align*}
    r_{p,s} = \frac{l_{p,s}}{\text{min}\{l_{p,s}: s\in \mathcal{S}\}}\,.
\end{align*}
From this we can define the performance profile as the probability for the algorithm $s \in \mathcal{S}$ that the performance ratio on any dataset is within a factor $\tau \in \mathbb{R}$ of the best possible ratio:
\begin{align*}
    \rho_s(\tau) = \frac{1}{27}\text{size}\{p \in \mathcal{P}: r_{p,s} \leq \tau\}\,,
\end{align*}
the function $\rho_s$ is the (cumulative) distribution function for the performance ratio. We compute the performance profile for each algorithm from the AutoML baseline, which can be found Figure~\ref{fig:perf_ratio}.
From the performance profile, we can see that compared to the baseline, DRAGON has an error close to the best for every dataset. It is also the only algorithm for which the performance ratio is less than two for all datasets. This diagram also suggests that the performance of AutoPytorch and AutoGluon are not that different.

\subsection{Computation time}

To be consistent with the other algorithms from the baseline, we set a fixed time budget of 8 hours for our experiments. But in most cases the algorithm found the best solution in less time than this. 
\begin{figure}[htbp]
    \centering
    \includegraphics[width=0.95\textwidth]{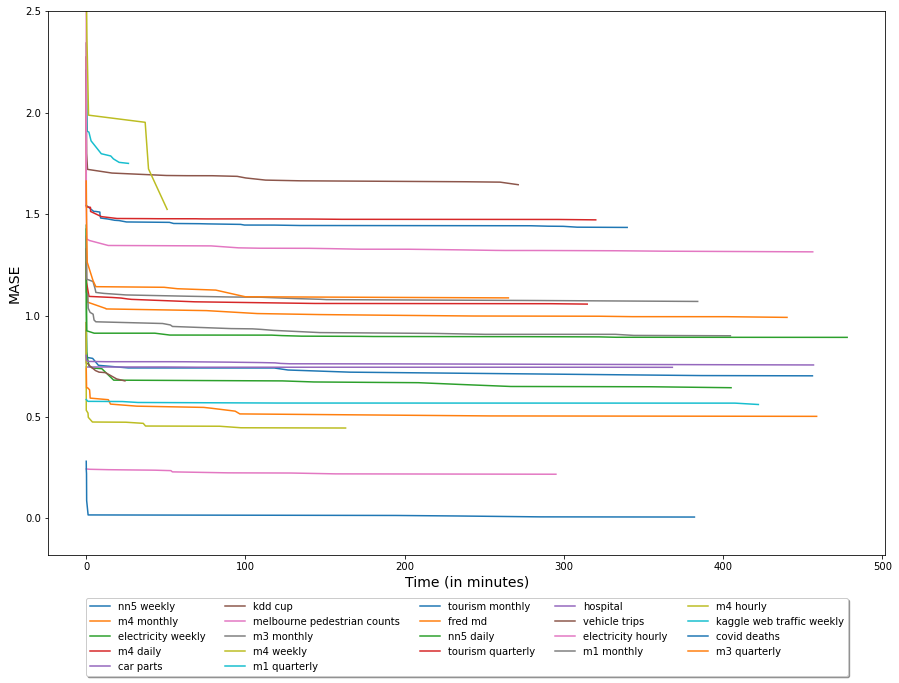}
    \caption{Computation time of DRAGON for each dataset. The curves represent the time when the best loss so far has been found for each dataset.}
    \label{fig:computation_time}
\end{figure}
Figure~\ref{fig:computation_time} represents the time convergence of DRAGON for each dataset. For almost every dataset, a close solution to the final one was found in less than an hour. For some datasets like M4 weekly or M1 Quarterly, DRAGON did not improve the results after the first hour. The models from the baseline train faster, with AutoGluon for example having an average runtime of 33 minutes \citep{shchur2023autogluon}. However, those algorithms are based on Machine Learning models, wherease the runtime of AutoPytorch, the other AutoDL framework was set to 4 hours for each datasets. The training time of DNNs are indeed usually higher than for traditional machine learning models. We think we can improve our training time using a multi-fidelity approach. Indeed, with our evolutionary algorithm, every DNN is trained until for 100 epochs before being evaluated. With a multi-fidelity approach we could speed up the identification of good performing models and stop training the worst ones sooner.

\subsection{Best models analysis}

In the AutoDL literature, few efforts are usually made to analyze the generated DNNs. In \citet{DBLP:journals/corr/abs-1909-09569} the authors established that architectures with wide and shallow cell structures are favoured by the NAS algorithms and that they suffer from poor generalization performance. We can rightfully ask ourselves about the efficiency of our framework and some of these questions may be answered thanks to a light models analysis. By the end of this section, we will try to answer some inquiries about our framework outputs. To answer those questions we defined some structural indicators, and we computed them in Table \ref{tab:best_models} for the best model for each dataset from \citet{godahewa2021monash}:

\begin{itemize}
    \item \textbf{\textit{Nodes}}: it represents the number of nodes (i.e. operations) in the graph.
    \item \textbf{\textit{Width}}: it represents the network width which can be defined as the maximum of incoming or outgoing edges to all nodes within the graph.
    \item \textbf{\textit{Depth}}: it defines the network depth which corresponds to the size of the longest path in the graph.
    \item \textbf{\textit{Edges}}: it represents the number of edges, relative to the number of nodes in the graph. It indicates how complex the graph can be and how sparse the adjacency matrix is.
    \item The last 7 indicators correspond to the number of the appearance of each layer type within the DNN.
\end{itemize}

\begin{table}[htbp]
\scriptsize
\begin{center}
\caption{\footnotesize{Structural indicators of the best model for each dataset found by DRAGON.}}
\label{tab:best_models}
\begin{tabular}{|c| c c c c c c c c c c c|}
\hline
Dataset & Nodes & Width & Depth & Edges & MLP & Att & CNN & RNN & Drop & Id & Pool \\
\hline
m3 monthly & 6 & 3 & 4 & 12 & 2 & 1 & 1 & 1 & 0 & 0 & 1 \\
covid death & 3 & 1 & 3 & 3 & 1 & 1 & 0 & 0 & 1 & 0 & 0 \\
m3 quarterly & 4 & 2 & 3 & 6 & 1 & 0 & 0 & 2 & 0 & 1 & 0 \\
vehicle trips & 4 & 3 & 4 & 8 & 1 & 0 & 0 & 1 & 0 & 0 & 2 \\
m1 yearly & 8 & 5 & 5 & 17 & 2 & 2 & 0 & 0 & 1 & 3 & 0 \\
m4 monthly & 6 & 3 & 4 & 12 & 2 & 1 & 1 & 1 & 0 & 0 & 1 \\
m3 other & 4 & 4 & 3 & 8 & 2 & 1 & 0 & 1 & 0 & 0 & 0 \\
tourism quarterly & 1 & 1 & 1 & 1 & 1 & 0 & 0 & 0 & 0 & 0 & 0 \\
pedestrian & 2 & 2 & 2 & 3 & 0 & 0 & 1 & 0 & 1 & 0 & 0 \\
nn5 daily & 5 & 5 & 3 & 12 & 1 & 1 & 1 & 0 & 0 & 2 & 0 \\
Web Traffic & 7 & 4 & 6 & 16 & 2 & 2 & 2 & 0 & 0 & 1 & 0 \\
m1 quarterly & 1 & 1 & 1 & 1 & 1 & 0 & 0 & 0 & 0 & 0 & 0 \\
tourism yearly & 7 & 3 & 6 & 15 & 2 & 1 & 1 & 0 & 1 & 1 & 1 \\
electricity weekly & 7 & 5 & 7 & 18 & 3 & 1 & 1 & 0 & 0 & 2 & 0 \\
m4 hourly & 5 & 4 & 5 & 11 & 0 & 2 & 1 & 0 & 1 & 1 & 0 \\
electricity hourly & 3 & 3 & 3 & 5 & 0 & 1 & 0 & 0 & 0 & 2 & 0 \\
m3 yearly & 6 & 4 & 5 & 13 & 1 & 2 & 0 & 0 & 0 & 3 & 0 \\
m4 weekly & 2 & 2 & 2 & 3 & 1 & 0 & 0 & 0 & 0 & 1 & 0 \\
m4 daily & 2 & 2 & 2 & 3 & 0 & 1 & 0 & 0 & 0 & 1 & 0 \\
nn5 weekly & 2 & 2 & 2 & 3 & 1 & 0 & 1 & 0 & 0 & 0 & 0 \\
kdd cup & 7 & 6 & 5 & 17 & 2 & 1 & 2 & 1 & 1 & 0 & 0 \\
hospital & 4 & 4 & 3 & 8 & 1 & 0 & 2 & 1 & 0 & 0 & 0 \\
m1 monthly & 2 & 1 & 2 & 2 & 1 & 0 & 0 & 0 & 0 & 1 & 0 \\
fred md & 5 & 4 & 3 & 9 & 0 & 1 & 1 & 0 & 1 & 1 & 1 \\
car parts & 7 & 3 & 6 & 15 & 2 & 1 & 1 & 0 & 1 & 1 & 1 \\
\hline
\hline
\textbf{Mean} & 4.40 & 3.08 & 3.60 & 8.84 & 1.20 & 0.80 & 0.64 & 0.32 & 0.32 & 0.84 & 0.28 \\
\hline
\end{tabular}
\end{center}
\end{table}

\textit{Does DRAGON always converge towards complex models, or is it able to find simple DNNs?}

From Table~\ref{tab:best_models} we notice that the model found are really small compare to a Transformer model for example and all have less than 8 hidden layers while we let the algorithms have cells with up to 10 nodes. Moreover, two models are made of only one layer. Another indicator of the models simplicity is the percentage of feed-forward and identity layers found in the best models. The feed-forward layer (also called MLP Table~\ref{tab:best_models}) is the most recurrent layer, as it appears on average at least once by graph, although our search space offers more complex layers such as convolution, recurrence or attention layers less frequently picked. This proves that even without regularization penalties, our algorithmic framework does not systematically search for over-complicated models. 

\vspace{0.5cm}
\textit{Does DRAGON always converge towards similar architectures for different datasets?}

\begin{figure}[htbp]
\centering

\begin{subfigure}[b]{0.35\columnwidth}
\centering
\includegraphics[width=\columnwidth]{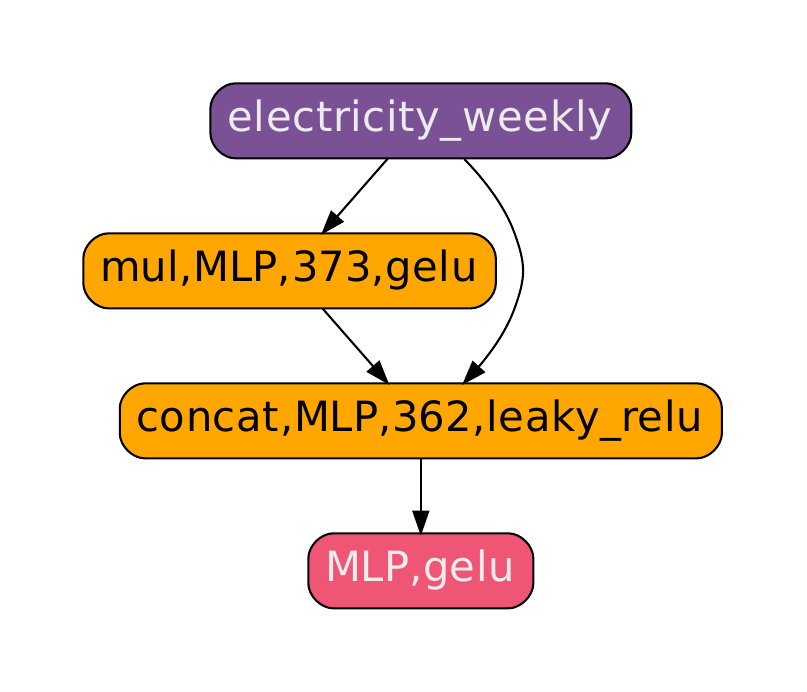}
\caption{MASE: 0.653}
\label{fig:simplegraph2}
\end{subfigure}
\begin{subfigure}[b]{0.64\columnwidth}
\centering
\includegraphics[width=\columnwidth]{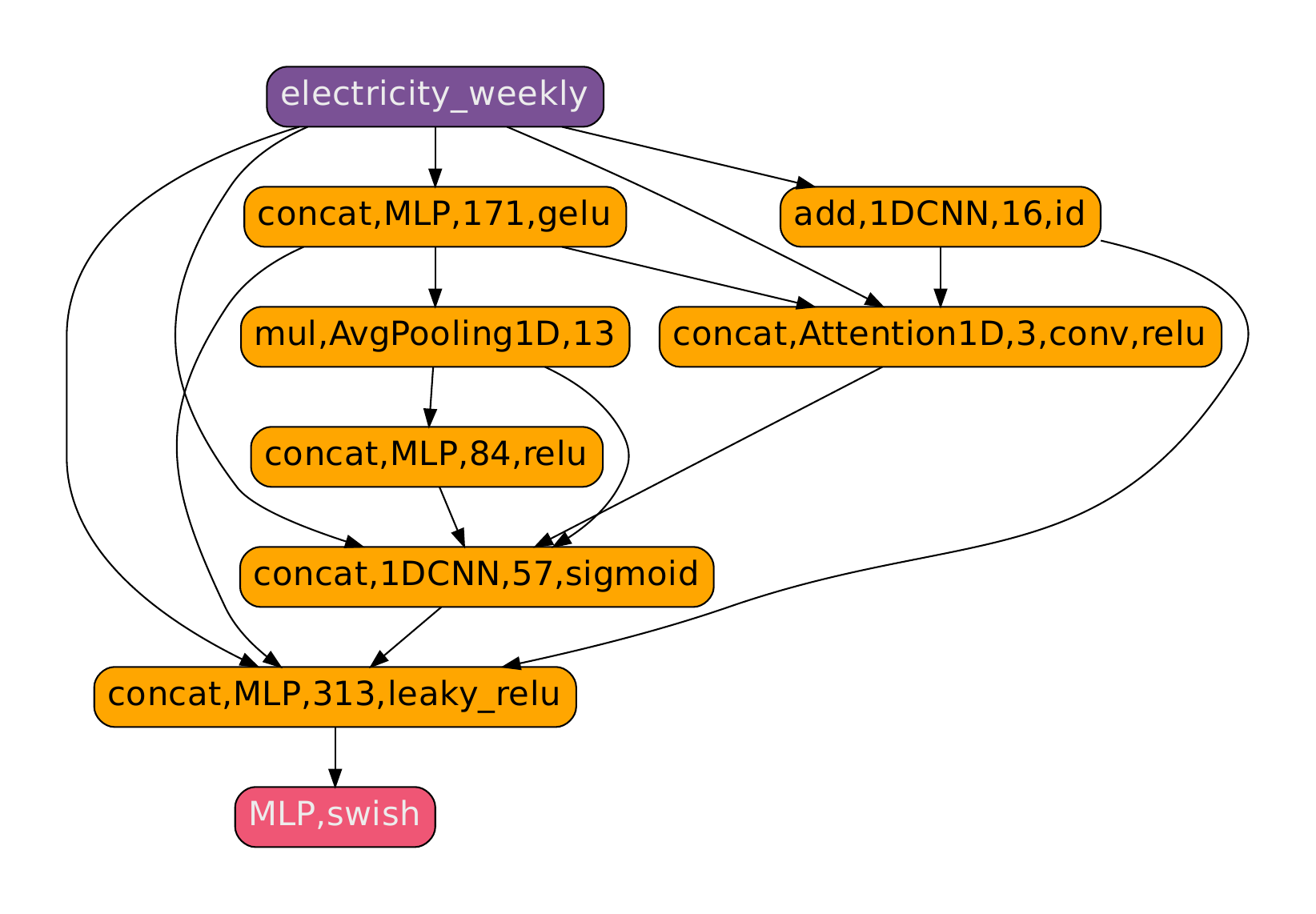}
\caption{MASE: 0.654}
\label{fig:complexgraph2}
\end{subfigure}

\caption{Two different models having similar good performance on the Electricity Weekly dataset (best MASE: 0.644).}
\label{fig:diff_graph_same_dataset}
\end{figure}

The structural indicators for all the datasets from Table~\ref{tab:best_models} are significantly different for each dataset, meaning the framework does not converge to similar architectures. As we set the seed, the initial population of size $K$ is identical for each dataset, but then the performance of the evaluated models affects the creation of the following graphs, which lead to different final model, optimized for each time series. Moreover, Figure~\ref{fig:diff_graph_same_dataset} shows that DRAGON may find very different performing architecture for a same dataset.

\vspace{0.5cm}
\textit{What is the diversity of the operations within the best models?}

The MLP layer is definitively the most used operation within the candidates ones. On average, each model from Table~\ref{tab:best_models} is using at least one MLP layer. Interestingly the CNN and Attention layers are more often used than RNN layers, which were designed for time series. Another intersting insight is that every candidate operation has been at least picked once, which states the operations diversity within the best models.

\vspace{0.5cm}
\textit{Are the best models still \say{deep} neural networks or are they wide and shallow as stated in \citet{DBLP:journals/corr/abs-1909-09569}?}

To answer this question, the observations from \citet{DBLP:journals/corr/abs-1909-09569} do not necessary apply to our results. Our models are on average a bit deeper than wide, bearing in mind that the indicators do not take into account the last MLP as shown Figure~\ref{fig:metamodel_monash}. If we were doing multi-fidelity in the future, this observation might change as one of the reasons mentioned in the paper for wider DNNs is the premature evaluation of architecture before full convergence.

\begin{figure}[htbp]
\centering

\begin{subfigure}[b]{0.25\columnwidth}
\centering
\includegraphics[height=\columnwidth]{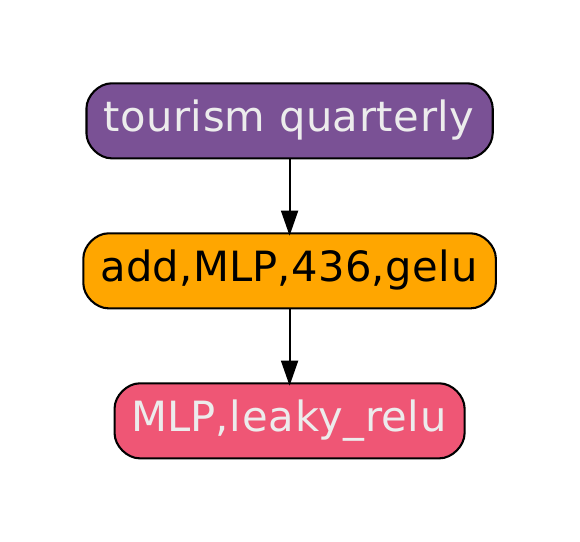}
\caption{Tourism Quarterly}
\label{fig:simplegraph}
\end{subfigure}
\begin{subfigure}[b]{0.74\columnwidth}
\centering
\includegraphics[width=\columnwidth]{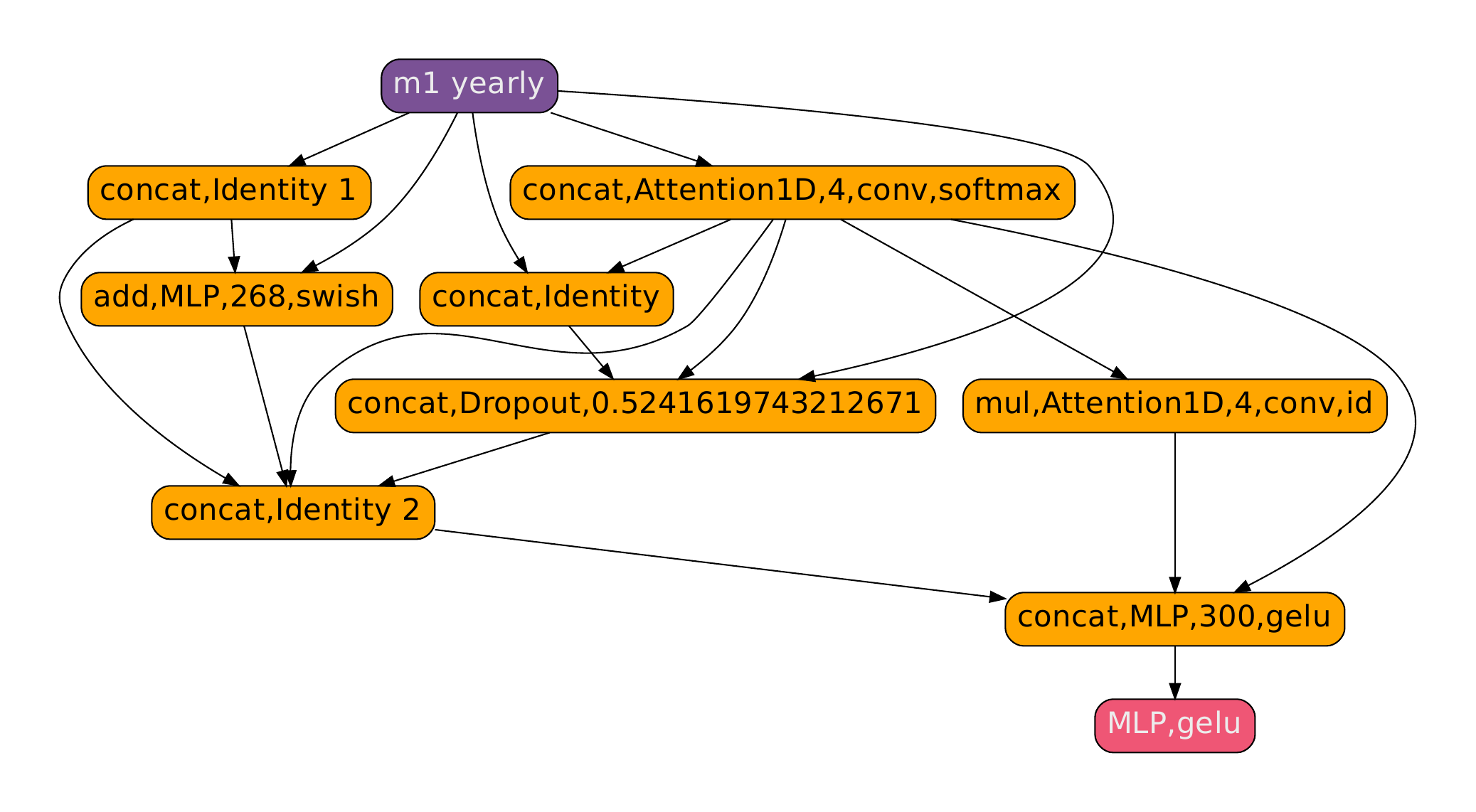}
\caption{M1 Yearly}
\label{fig:complexgraph}
\end{subfigure}

\begin{subfigure}[b]{0.52\columnwidth}
\centering
\includegraphics[width=\columnwidth]{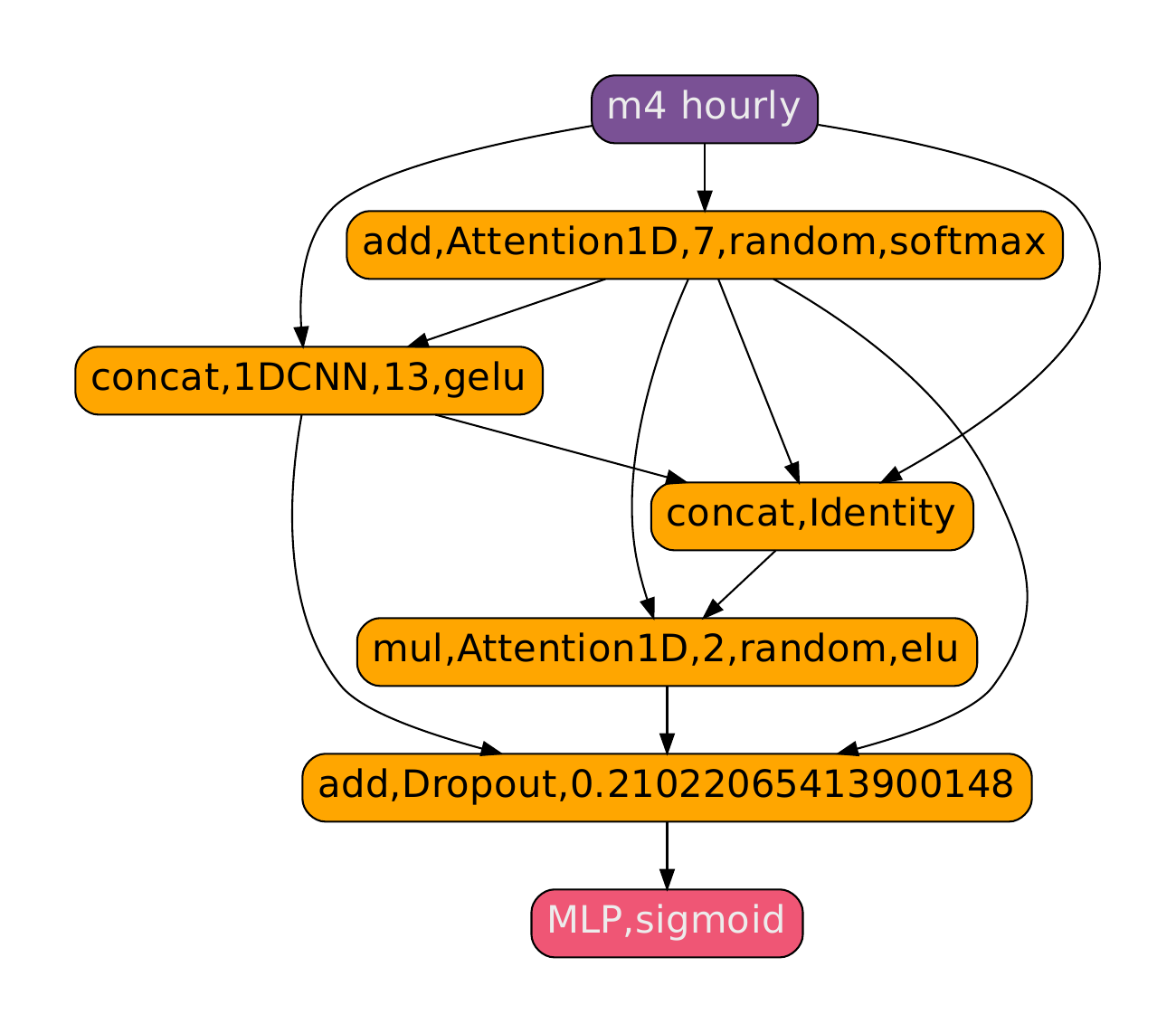}
\caption{M4 Hourly}
\label{fig:m4_hourly}
\end{subfigure}
\begin{subfigure}[b]{0.45\columnwidth}
\centering
\includegraphics[width=\columnwidth]{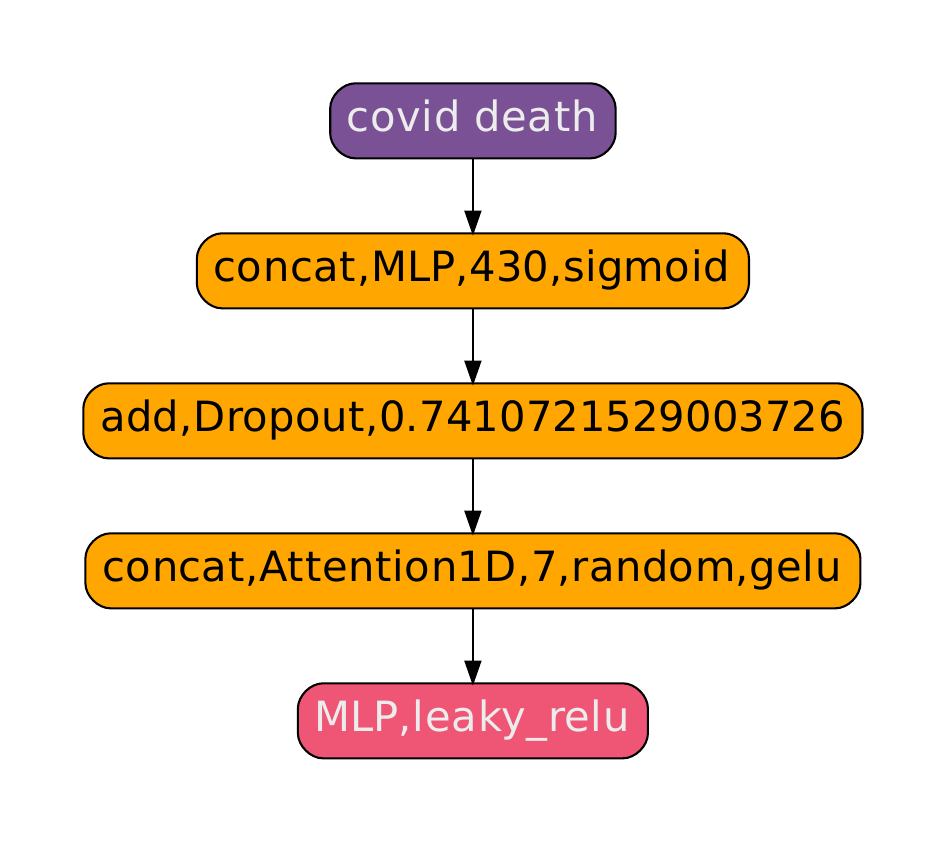}
\caption{Covid Death}
\label{fig:covid_deaths}
\end{subfigure}

\caption{Best DNNs output by our algorithmic framework.}
\label{fig:diverse_graphs}
\end{figure}

\subsection{Ablation study}

We chose two datasets, M1 monthly and Tourism monthly, in order to reduce the number of experiments we had to perform, as the benchmark was quite large. We conducted tests for four search algorithms for each dataset; these were random search, a population based evolutionary algorithm (EA) with alternating optimisation of hyperparameters and architecture, as well as a version with joint optimisation and, lastly, simulated annealing. To explore the search space, we used an exponential multiplicative monotonic cooling schedule in our simulated annealing algorithm: $T_k = T_0.\alpha^k$. We evaluated 40 neighborhood solutions at each iteration to accomplish this. To ensure fairness between each search algorithm, we conducted five experiments with different seeds (0, 100, 200, 300, and 400), and parameterised the algorithms to evaluate 4000 DNNs. The results of this study are presented in Table \ref{tab:ablation}.

\newcolumntype{Y}{>{\centering\arraybackslash}X}
\begin{table}[htbp]
\caption{Comparison between several search algorithms over two datasets: M1 Monthly and Tourism Monthly. Each configuration has been ran with five different seeds.}
\begin{center}
\begin{tabularx}{\linewidth}{|Y|Y Y|}
        \hline
        \textbf{Search Algorithm} & M1 Monthly & Tourism Monthly \\
        \hline
        \hline
        Random search & $1.098 \pm 0.006$ & $1.645 \pm 0.018$ \\
        \hline 
        EA joint mutation & $\boldsymbol{1.073 \pm 0.004}$  & $\boldsymbol{1.450 \pm 0.003}$\\
        \hline
        EA alternating mutation & $1.080 \pm 0.005$ & $1.451 \pm 0.004$\\
        \hline
        Simulated Annealing & $1.141 \pm 0.044 $& $2.640 \pm 0.037$\\
        \hline                       
    \end{tabularx}
\label{tab:ablation}
\end{center}
\end{table}

\begin{figure}[htbp]
\centering
\begin{subfigure}[b]{0.49\columnwidth}
\centering
\includegraphics[width=\columnwidth]{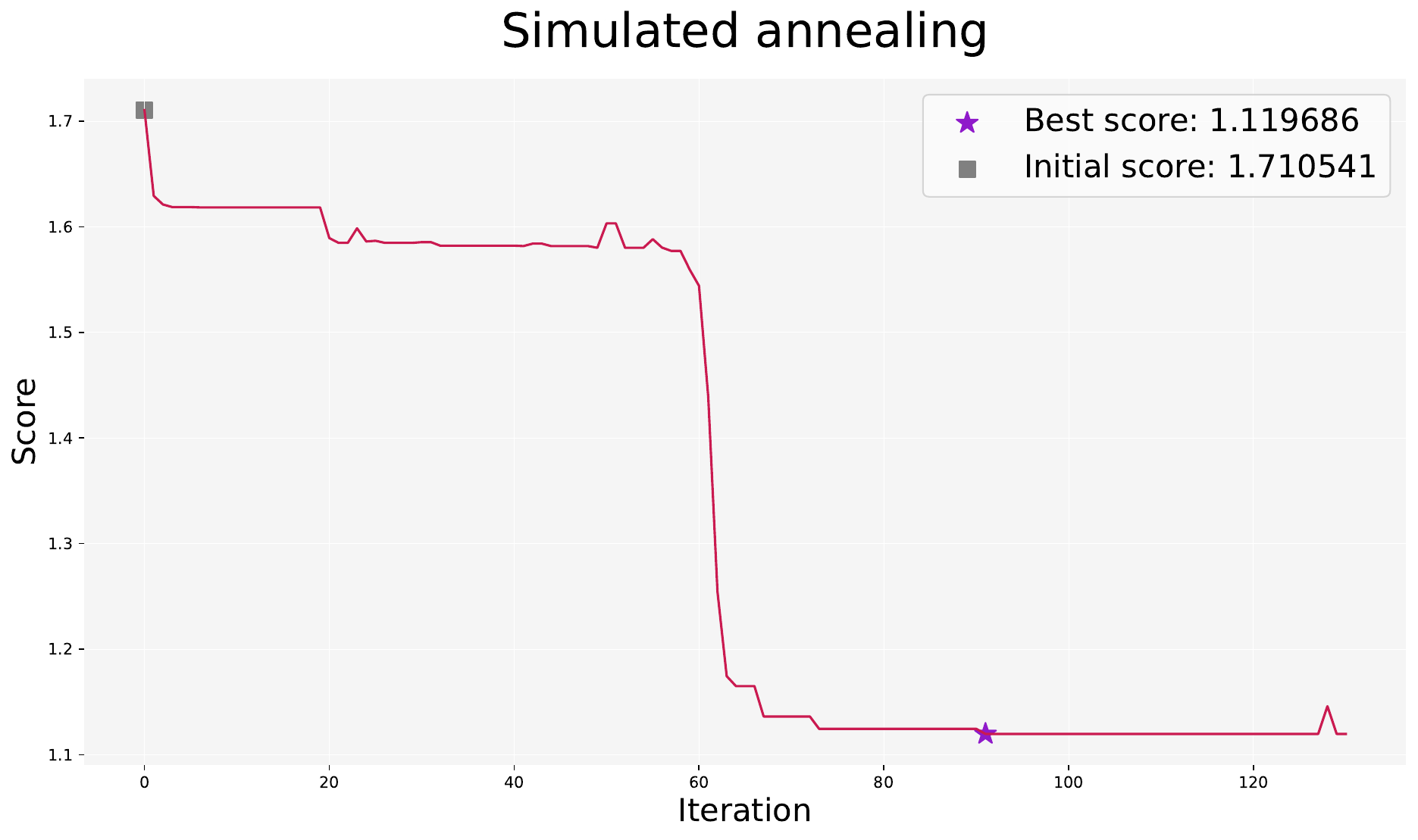}
\caption{\centering Best score for each iteration of the simulated annealing algorithm.}
\label{fig:lineplot_sa}
\end{subfigure}
\begin{subfigure}[b]{0.49\columnwidth}
\centering
\includegraphics[width=\columnwidth]{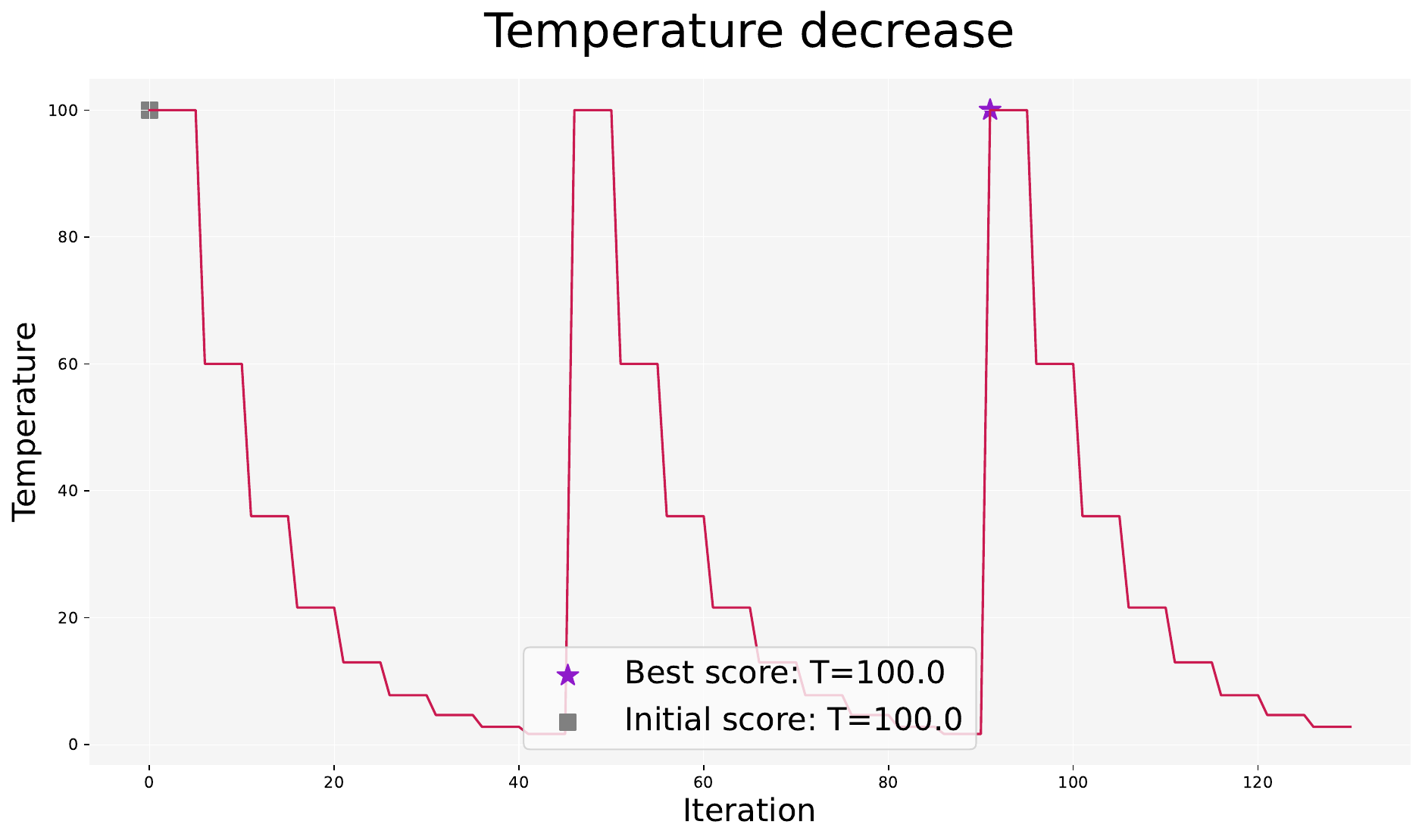}
\caption{\centering Temperature value for each iteration of the simulated annealing algorithm.}
\label{fig:templot_sac}
\end{subfigure}

\caption{\centering Simulated annealing algorithm for the M1 Monthly dataset, with seed=100. MASE=1.120.}
\label{fig:sa}
\end{figure}

The findings suggest that the most exploratory algorithms, specifically the random search algorithm and the evolutionary algorithm, yielded better results than the more locally focused, ie: the simulated annealing. The findings imply the presence of several potential solutions in the search space, but none of them could be accessed by the simulated annealing algorithms from their starting points. The figure \ref{fig:sa} indicates that the most effective DNN was achieved with the help of simulated annealing when $Temp=Temp_{max}$. This suggests greater exploration by the algorithm. Additionally, the random search method produced favorable results. The assessment of 4000 solutions for the Tourism Monthly dataset and M1 dataset was completed within 12 minutes and 4 hours respectively, thanks to the parallelisation of the solution. This shows that the search space has been suitably designed for our problem. However, it does not achieve the same level of performance as our evolutionary algorithms, highlighting the significance of our variational operators. Ultimately, both types of mutation produce very similar results.

\subsection{Nondeterminism and instability of DNNs}

\begin{figure}[htbp]
\centering

\begin{subfigure}[b]{0.49\columnwidth}
\centering
\includegraphics[width=\columnwidth]{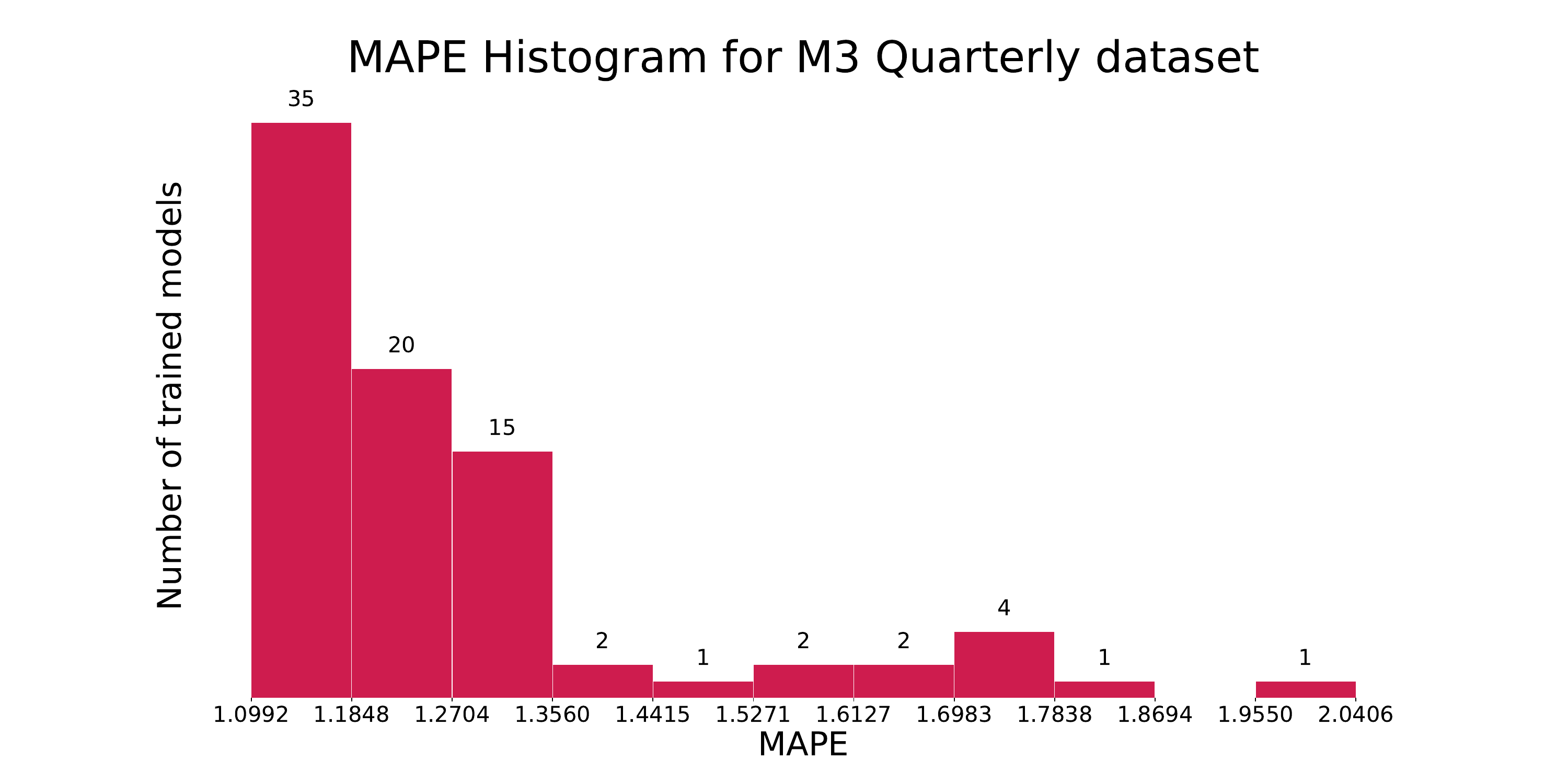}
\caption{M3 Quarterly, best MASE: 1.099}
\label{fig:seeds_m3}
\end{subfigure}
\begin{subfigure}[b]{0.49\columnwidth}
\centering
\includegraphics[width=\columnwidth]{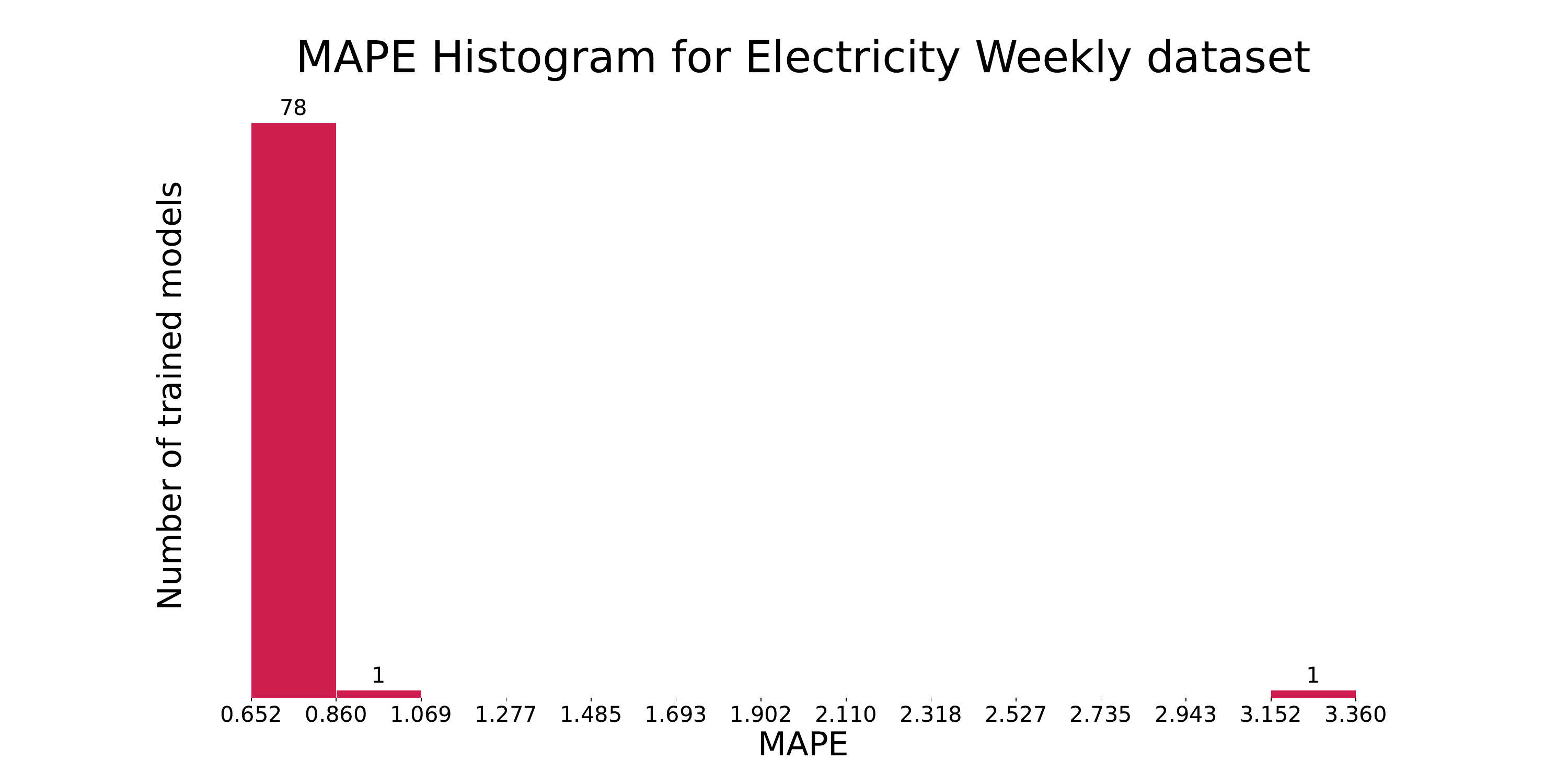}
\caption{Electricity Weekly, best MASE: 0.652}
\label{fig:seeds_elec}
\end{subfigure}

\caption{MASE histogram of the best model performances with multiple seeds for two datasets.}
\label{fig:seeds_histo}
\end{figure}

An often overlooked robustness challenge with DNN optimization is their uncertainty in performance \citep{summers2021nondeterminism}. A unique model with a fixed architecture and set of hyperparameters can produce a large variety of results on a dataset. Figure \ref{fig:seeds_histo} shows the results on two datasets: M3 Quarterly and Electricity Weekly. For both datasets, we selected the best models found with our optimization and drew 80 seeds summing all instability and nondeterministic aspects of our models. We trained these models and plotted the MASE Figure \ref{fig:seeds_histo}. On the M3 Quarterly, the MASE reached values two times bigger than our best result. On the Electricity Weekly, it went up to five times worst. To overcome this problem, we represented the parametrization of stochastic aspects in our models as a hyperparameter, which we added to our search space. Despite its impact on the performance, we have not seen any work on NAS, HPO or AutoML trying to optimize the seed of DNNs. Our plots of Figure \ref{fig:seeds_histo} showed that the optimization was effective as no other seeds gave better results than the one picked by DRAGON.

\section{Conclusion and Future Work} \label{part 6}

In this article, we introduce DRAGON, a novel algorithmic framework to optimize jointly the architectures of DNNs and their hyperparameters. We initially presented a search space founded on Directed Acyclic Graphs, which is flexible for architecture optimization and also allows fine-tuning of hyperparameters. We then develop search operators that are compatible with any metaheuristic capable of handling a mixed and variable-size search space. We prove the efficiency of DRAGON on a task rarely tackled by AutoDL or NAS works: time series forecasting. On this task where the performing DNNs have not been clearly identified, DRAGON shows superior forecasting capalities compare to the state-of-the-art in AutoML and handcrafted models. 

Although we obtained satisfactory results compared to our baseline, we note that our algorithm runs slower than AutoGluon, its main competitor, and does not improve it much. However, we would like to point out that AutoGluon produces mixtures of machine learning models, while DRAGON produces a single DNN. To be more competitive in terms of computation time and results, we could consider using multi-fidelity techniques to identify and eliminate unpromising solutions more quickly, using multi-objective techniques to increase the value of simpler, easier-to-train DNNs, and taking inspiration from AutoGluon and AutoPytorch techniques and blending DNNs and machine learning predictors to further improve forecasting accuracy. Moreover, for each generated architecture, we optimize the hyperparameters using the same evolutionary algorithm. However, hyperparameters play a large role in the performance of a given architecture, and it could be interesting to investigate an optimization that alternates between specific search algorithms for the architecture and for the hyperparameters. In fact, while the graph structure representing the architecture is difficult to manipulate, once fixed, the hyperparameter search space can be considered as a vector that could be optimized with more efficient algorithms such as Bayesian or bi-level optimization, allowing a greater number of possibilities to be evaluated.

Furthermore, given our search space and search algorithms' universality, we could extend our framework to several other tasks. Indeed, only the candidate operations included as node content are task-related, and the representation of DNNs as DAG is not. Further research can test our framework on various learning tasks, necessitating the creation of new operations, such as 2-dimensional convolution and pooling, for the treatment of images, for example. Additionally, this framework can also function as a cell-based search space, utilising normal and reduction cells as opposed to a single convolution operation. 

Finally, our study demonstrates that incorporating a variety of cutting-edge DNN operations into a single model presents a promising approach for enhancing the performance of time series forecasting. We consider these models as innovative within the deep learning community, and further research investigating their efficacy could be interested.


\acks{We would like to thank Margaux Brégère for her careful proofreading and helpful advice. This work has been funded by Electricité de France (EDF). The supercomputers used to run the experiments belong to EDF.}


\newpage

\appendix
\section{Available operations and hyperparameters}
\label{anx:op_hp}
\begin{table}[h!]
\caption{Operations available in our search space and used for the Monash time series archive dataset and their hyperparameters that can be optimized.}
\begin{center}
\begin{tabularx}{\linewidth}{|Y|Y Y|}
        \hline
        \textbf{Operation} & \multicolumn{2}{c|}{\textbf{Optimized hyperparameters}} \\
        \hline
        \hline
        Identity & \multicolumn{2}{c|}{-} \\
        \hline
        Fully-Connected (MLP) & Output shape & Integer\\
        \hline
        \multirow{2}{*}{Attention}
                       & Initialization type & [convolution, random] \\
                       & Heads number & Integer \\
        \hline
        1D Convolution & Kernel size & Integer \\
        \hline
        \multirow{2}{*}{Recurrence}
                        & Output shape & Integer\\
                       & Recurrence type & [LSTM, GRU, RNN] \\
        \hline
        \multirow{2}{*}{Pooling}
                        & Pooling size & Integer\\
                       & Pooling type & [Max, Average] \\
        \hline
        Dropout & Dropout Rate & Float\\      
        \hline                       
    \end{tabularx}
\label{tab1}
\end{center}
\end{table}

Activation functions, $\forall x \in \mathbb{R^D}$
\begin{itemize}
    \item Id: $\mathrm{id}(x) = x$
    \item Sigmoid: $\mathrm{sigmoid}(x)={\frac {1}{1+{\rm {e}}^{-x}}} $
    \item Swish: $\mathrm{swish}(x) = x \times \mathrm{sigmoid}(\beta x) ={\frac {x}{1+e^{-\beta x}}}$
    \item Relu: $\mathrm{relu}(x) = \max(0,x)$
    \item Leaky-relu: $\mathrm{leakyRelu}(x) = \mathrm{relu}(x) + \alpha \times \min(0,x)$, in our case: $\alpha = 10^{-2}$
    \item Elu: $\mathrm{elu}(x) = \mathrm{relu}(x) + \alpha \times \min(0, e^x - 1)$
    \item Gelu: $\mathrm{gelu}(x) = x\mathbb{P}(X\leq x) \approx 0.5x(1+\tanh[\sqrt{2/\pi}(x+0.044715x^3)])$
    \item Softmax: $\sigma(\mathbf{x})_j=\frac{\mathrm e^{x_j}}{\sum_{d=1}^D\mathrm e^{x_d}}$ $\forall j\in\left\{1,\ldots,D \right\}$
\end{itemize}

\newpage

\section{Monash datasets presentation} \label{part info monash}
\newcolumntype{Y}{>{\centering\arraybackslash}X}

\begin{table}[h!]
\scriptsize
\begin{center}
\caption{Information about the Monash datasets \citep{godahewa2021monash}.}
\label{tab:info}
\begin{tabularx}{\linewidth}{|Y|Y Y Y Y Y|}
\hline

\textbf{Dataset} & \textbf{Domain} & \textbf{Nb of series} & \textbf{Multivariate} & \textbf{Lag} & \textbf{Horizon} \\
\hline
\hline
Carparts & Sales & 2674 & Yes & 15 & 12 \\
\hline
Elec. hourly & Energy & 321 & Yes & 30 & 168 \\
\hline
Elec. weekly & Energy & 321 & Yes & 65 & 8 \\
\hline
Fred MD & Economic & 107 & Yes & 15 & 12 \\
\hline
Hospital & Health & 767 & Yes & 15 & 12 \\
\hline
KDD & Nature & 270 & No & 210 & 168 \\
\hline
\makecell{M1\\monthly} & Multiple & 1001 & No & 15 & 18 \\
\hline
M1 quart.  & Multiple & 1001 & No & 5 & 8 \\
\hline
M1 yearly & Multiple & 1001 & No & 2 & 6 \\
\hline
\makecell{M3\\monthly}  & Multiple & 3003 & No & 15 & 18 \\
\hline
M3 other & Multiple & 3003 & No & 2 & 8 \\
\hline
M3 quart. & Multiple & 3003 & No & 5 & 8 \\
\hline
M3 yearly & Multiple & 3003 & No & 2 & 6 \\
\hline
M4 daily & Multiple & 100000 & No & 9 & 14 \\
\hline
M4 hourly & Multiple & 100000 & No & 210 & 48 \\
\hline
\makecell{M4\\monthly}& Multiple & 100000 & No & 15 & 18 \\
\hline
M4 quart. & Multiple & 100000 & No & 5 & 8 \\
\hline
M4 weekly & Multiple & 100000 & No & 65 & 13 \\
\hline
NN5 daily & Banking & 111 & Yes & 9 & 56 \\
\hline
NN5 weekly & Banking & 111 & Yes & 65 & 8 \\
\hline
Pedestrians & Transport & 66 & No & 210 & 24\\
\hline
\makecell{Tourism\\monthly} & Tourism & 1311 & No & 2 & 24 \\
\hline
\makecell{Tourism\\quart.} & Tourism & 1311 & No & 5 & 8 \\
\hline
\makecell{Tourism\\yearly }& Tourism & 1311 & No & 2 & 4 \\
\hline
\makecell{Traffic\\weekly} & Transport & 862 & Yes & 65 & 8 \\
\hline
\makecell{Vehicle\\trips} & Transport & 329 & No & 9 & 30 \\
\hline
\end{tabularx}
\end{center}
\end{table}

\vskip 0.2in

\bibliography{my_bib}

\end{document}